\documentclass[10pt,twocolumn,letterpaper]{article}

\usepackage{iccv}
\usepackage{times}
\usepackage{epsfig}
\usepackage{graphicx}
\usepackage{amsmath}
\usepackage{amssymb}
\usepackage{subfig}
\usepackage{algorithm}
\usepackage{algorithmic}
\usepackage{tabularx}
\usepackage{enumitem}
\usepackage{amsmath}
\usepackage{booktabs}
\usepackage{makecell}

\usepackage[pagebackref=true,breaklinks=true,letterpaper=true,colorlinks,bookmarks=false]{hyperref}

\renewcommand{\thefootnote}{*}

\iccvfinalcopy 

\def\iccvPaperID{2452} 

\ificcvfinal\fi
\begin{document}

\newcommand{\Commentl}[1]{ {\hfill $\triangleright$ \it #1}}
\newcommand{\Comment}[1]{ {\begin{flushright} $\triangleright$ \it #1 \end{flushright}}}

\title{ \vspace{-0.2in} AlphaX: eXploring Neural Architectures with Deep Neural Networks \\ and Monte Carlo Tree Search \vspace{-0.2in}}

\author{Linnan Wang$^1$ \quad Yiyang Zhao \quad Yuu Jinnai$^1$  \quad Yuandong Tian$^2$ \quad Rodrigo Fonseca$^1$ \\
$^1$Brown University \quad $^2$Facebook AI Research}


\maketitle
\renewcommand{\thefootnote}{\arabic{footnote}}

\begin{abstract}
We present AlphaX, a fully automated agent that designs complex neural architectures from scratch.
AlphaX explores the search space with a distributed Monte Carlo Tree Search (MCTS) and a Meta-Deep Neural Network (DNN). MCTS guides transfer learning and intrinsically improves the search efficiency by dynamically balancing the exploration and exploitation at fine-grained states, while Meta-DNN predicts the network accuracy to guide the search, and to provide an estimated reward to speed up the rollout. As the search progresses, AlphaX also generates the training data for Meta-DNN. So, the learning of Meta-DNN is end-to-end. In 8 GPU days, AlphaX found an architecture that reaches 97.88\% top-1 accuracy on CIFAR-10, and 75.5\% top-1 accuracy on ImageNet. We also evaluate AlphaX on a large scale NAS dataset for reproducibility. On NASBench-101, AlphaX also demonstrates 3x and 2.8x speedup over \textit{Random Search} and \textit{Regularized Evolution} in finding the global optimum. Finally, we show the searched architecture improves a variety of vision applications from Neural Style Transfer, to Image Captioning and Object Detection.
\end{abstract}

\section{Introduction}

Designing efficient neural architectures is extremely laborious. A typical design iteration starts with a heuristic design hypothesis from domain experts, followed by the design validation with hours of GPU training. The entire design process renders many of such iterations before finding a satisfying architecture. Neural Architecture Search (NAS) has emerged as a promising tool to alleviate human effort in this trial and error design process, but the tremendous computing resources required by current NAS methods motivates us to investigate the search efficiency.

\begin{figure}[t]
\centering 
\subfloat[][random search]{\includegraphics[height = 3cm]{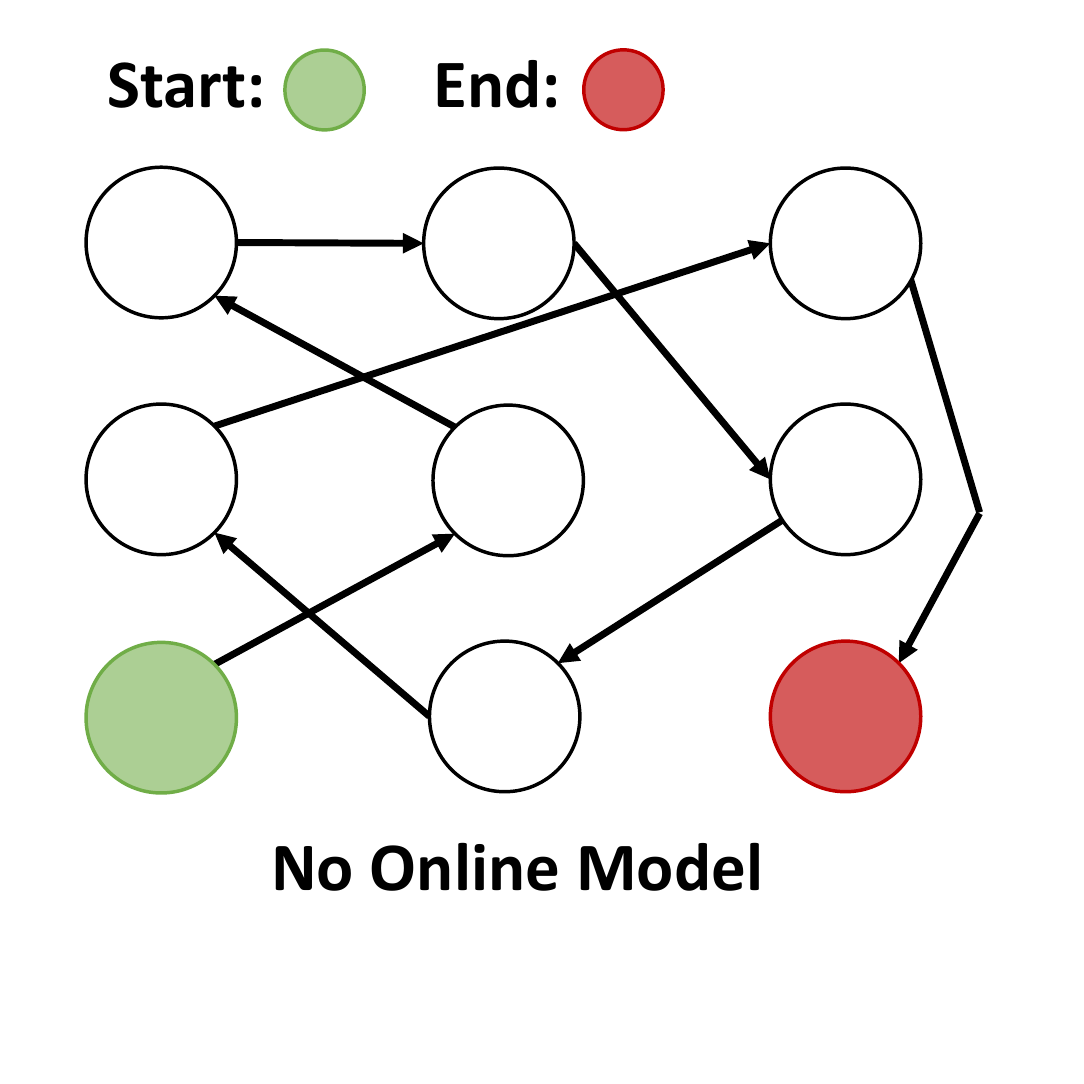}\label{design_domain}} \quad
\subfloat[][greedy methods]{\includegraphics[height = 3cm]{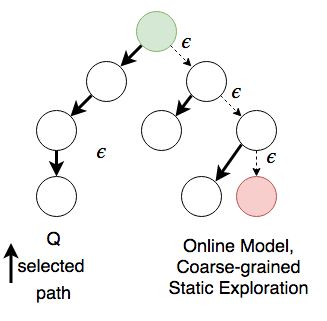}\label{MCTS_terminal_state}}\\
\subfloat[][MCTS]{\includegraphics[height = 3cm]{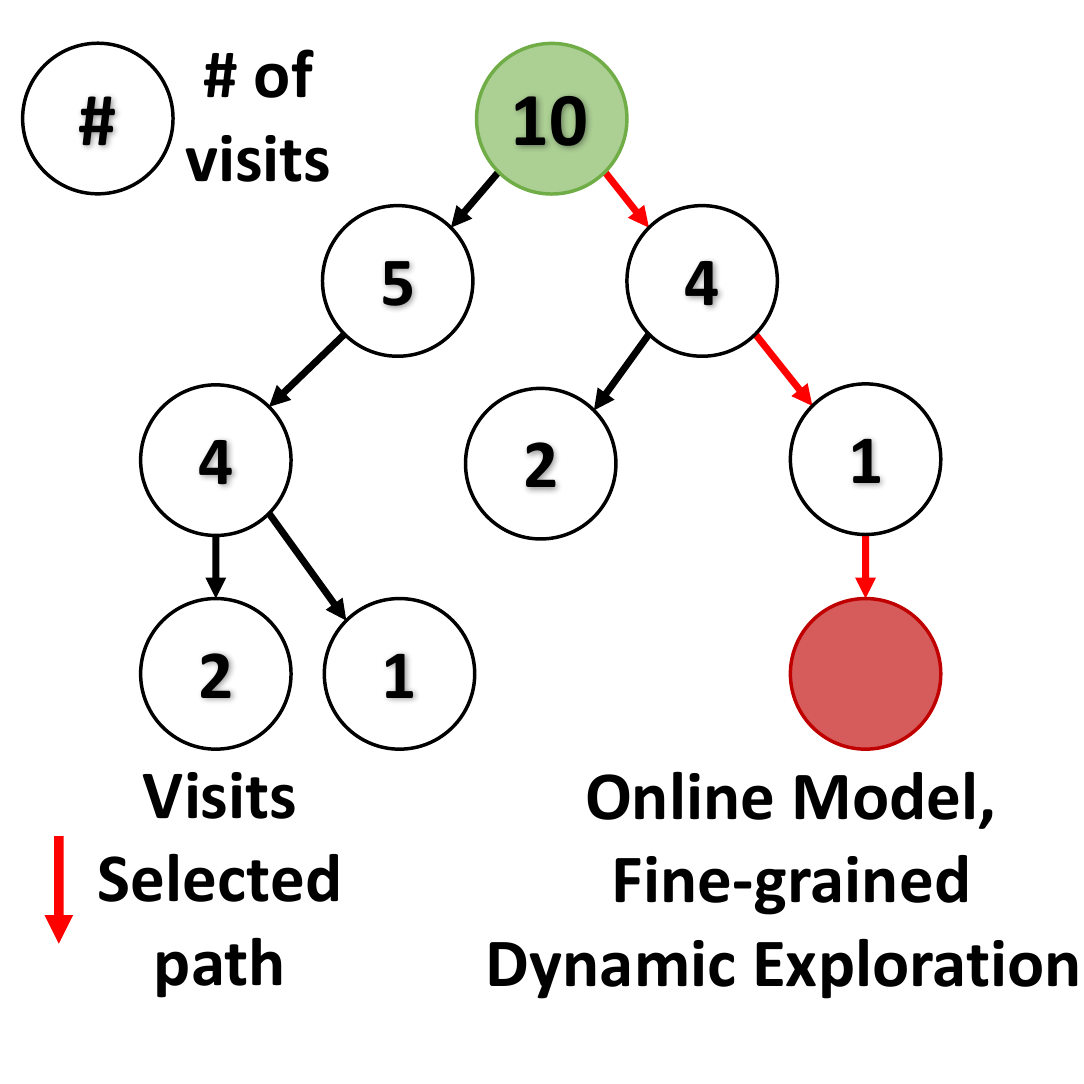}\label{MCTS_terminal_state}} \quad
\subfloat[][NASBench Performance]{\includegraphics[height = 3cm]{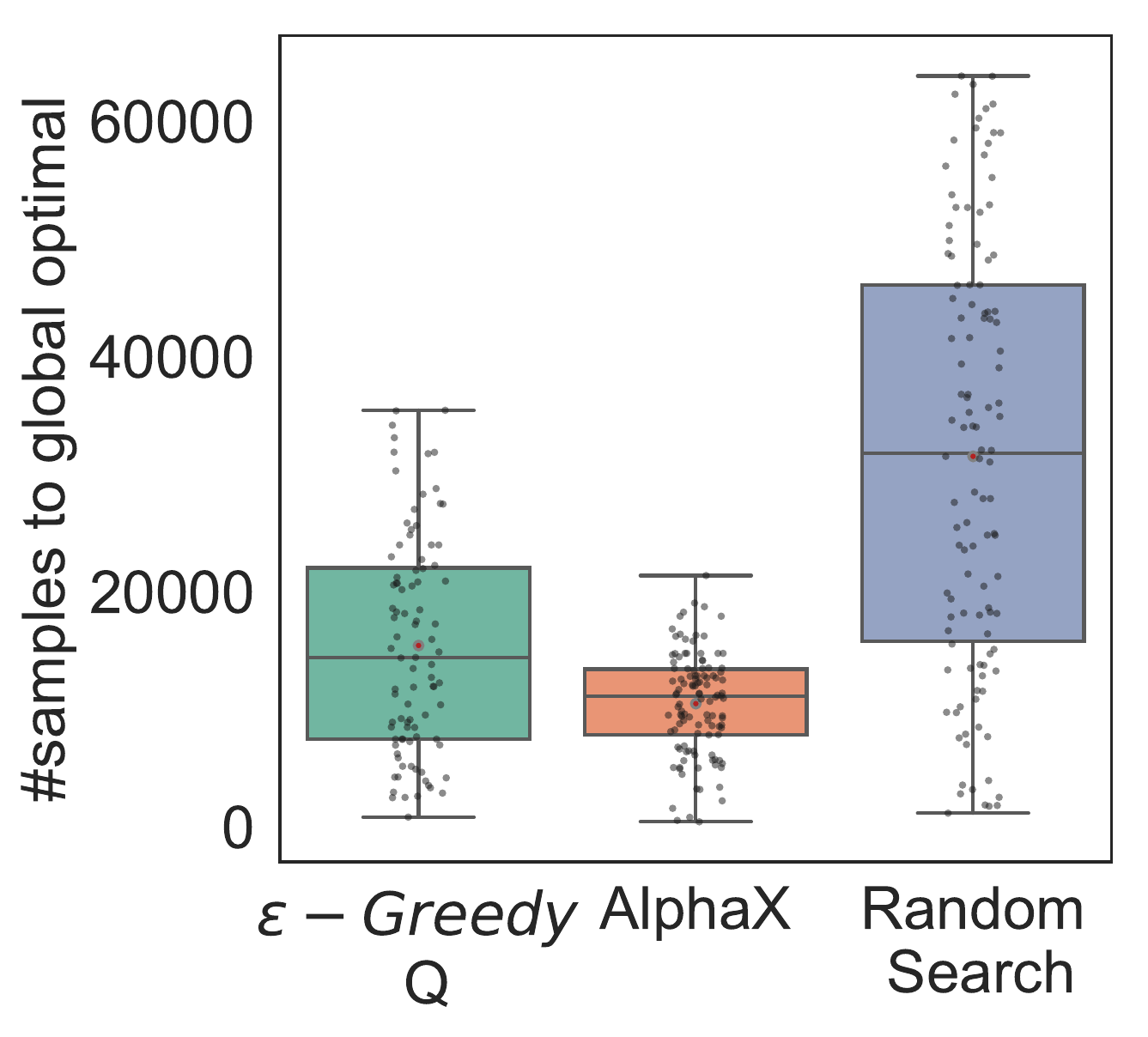}\label{mcts_speed_nasbench_teaser}}
    \caption{ \textbf{Comparisons of NAS algorithms}: \textbf{(a)} \textit{random search} makes independent decision without using prior rollouts (previous search trajectories). An online model is to evaluate how promising the current search branch based on prior rollouts, and Random Search has no online model. \textbf{(b)} Search methods guided by online performance models built from previous rollouts. With static, coarse-grained exploration strategy (e.g., $\epsilon$-greedy in \textit{Q-learning}), they may quickly be stuck in a sub-optimal solution; and the chance to escape is exponentially decreasing along the trajectory.  \textbf{(c)} AlphaX builds online models of both performance and visitation counts for adaptive exploration. \textbf{(d)} Performance of different search algorithms on NASBench-101 \cite{ying2019bench}. AlphaX is 3x, 1.5x more sample-efficient than \textit{random search} and \textit{$\epsilon$-greedy} based Q learning.  }
\label{fig:teaser}
\end{figure}

AlphaGo/AlphaGoZero~\cite{silver2016mastering} recently show super-human performance in playing the game of Go, by using a specific search algorithm called Monte-Carlo Tree Search (MCTS)\cite{kocsis2006bandit,browne2012survey}. Given the current game state, MCTS gradually builds an online model for its subsequent game states to evaluate the winning chance at that state, based on search experiences in the current and prior games, and makes a decision. Search experience is from the previous search trajectories (called \emph{rollouts}) that have been tried, and their consequences (whether the player wins or not). Different from traditional MCTS approach that evaluates the consequence of a trajectory by random self-play to the end of a game, AlphaGo uses a \emph{predictive model} (or value network) to predict the consequence, which enjoys much lower variance. Furthermore, due to its built-in exploration mechanism using \textit{Upper Confidence bound applied to Trees} (UCT)\cite{auer2002finite}, based on its online model, MCTS dynamically adapts itself to the most promising search regions, where good consequences are likely to happen.

Inspired by this idea, we present AlphaX that uses MCTS for efficient model architecture search with Meta-DNN as a predictive model to estimate the accuracy of a sampled architecture. Compared with Random Search, AlphaX builds an online model which guides the future search, compared to greedy methods, e.g. Q-learning, Regularized Evolution or Top-K methods, AlphaX dynamically trades off exploration and exploitation and can escape from locally optimal solutions with fewer number of search trials. Fig.~\ref{fig:teaser} summarizes the trade-offs. Furthermore, while prior works on MCTS applied to Architecture Search~\cite{wistuba2017finding, negrinho2017deeparchitect} only report performance on CIFAR-10, our distributed AlphaX system can scale up to 32 machines (with 32 acceleration ratio), reports better accuracies on CIFAR-10 and achieves performance on large-scale dataset like ImageNet that is on par with the SOTA. Particularly, AlphaX is up to 3x faster than Random Search and Regularized Evolution in finding the best performing network on NASBench-101.



\begin{table}[t]
\setlength{\tabcolsep}{0.2em}
  \scriptsize
  \label{layers_used}
  \centering
  \hspace{-1cm}
  \begin{tabular}{ l l l l l l }
    \toprule
          \textbf{Methods}     &\textbf{\thead{\scriptsize Global\\\scriptsize Solution }}     & \textbf{\thead{\scriptsize Online\\\scriptsize Model }}     & \textbf{Exploration} & \textbf{\thead{\scriptsize Distributed\\\scriptsize Ready }} & \textbf{\thead{\scriptsize Transfer\\\scriptsize Learning }} \\
    \midrule
		  MetaQNN (QL) \cite{Baker2016}               		             &  $\times$                     & -                 & {$\epsilon$-greedy}    & $\times$ & $\times$ \\
		  Zoph (PG)\cite{Zoph2016}                 		                 &  $\times$                     & RNN                      & -                      & $\surd$  & $\times$ \\
		  PNAS (HC)\cite{liu2017progressive}      			             &  $\times$                     & RNN                      & -                      & $\surd$  & $\times$ \\
		  Regularized Evolution\cite{real2018regularized}      	         &  $\surd$                      & -                        & top-k mutation         & $\surd$  & $\times$ \\
		  Random Search\cite{sciuto2019evaluating}                       &  $\surd$                      & -                        & random                 & $\surd$  & $\times$ \\
  		  DeepArchitect (MCTS)\cite{negrinho2017deeparchitect}    		 &  $\surd$					   & -                  & UCT                    & $\times$ & $\times$ \\	
		  Wistuba (MCTS)\cite{wistuba2017finding}            		     &  $\surd$                      & \thead{\scriptsize{Gaussian}}             & UCT      & $\times$ & $\times$ \\   
		  AlphaX (MCTS)                       &  $\surd$                      & meta-DNN                 & UCT           & $\surd$  & $\surd$  \\		  
    \bottomrule
    \label{acc-comps-imagenet}
  \end{tabular}
  \caption{\textbf{Highlights of NAS search algorithms}: compared with DeepArchitect and Wistuba, AlphaX features a scalable online model, and is ready for distributed system.}
  
  \vspace{-0.2in}
\end{table}

\section{Related Work}

\textit{Monte Carlo Tree Search:} DeepArchitect\cite{negrinho2017deeparchitect} implemented vanilla MCTS for NAS without a predictive model, and Wistuba \cite{wistuba2017finding} uses statistics from the current search (e.g., RAVE and Contextual Reward Prediction) to predict performance of a state. In comparison, our performance estimate is from both searched rollouts so far and a model (meta-DNN) learned from performances of known architectures and can generalize to unseen architectures. Both previous works report performance on CIFAR-10, while AlphaX reports performance on CIFAR-10, ImageNet, and NASBench-101. Most importantly, AlphaX is the first scalable MCTS based design agent.

\begin{figure}
\vspace{-0.2in}
\centering 
\includegraphics[width=1.1\columnwidth]{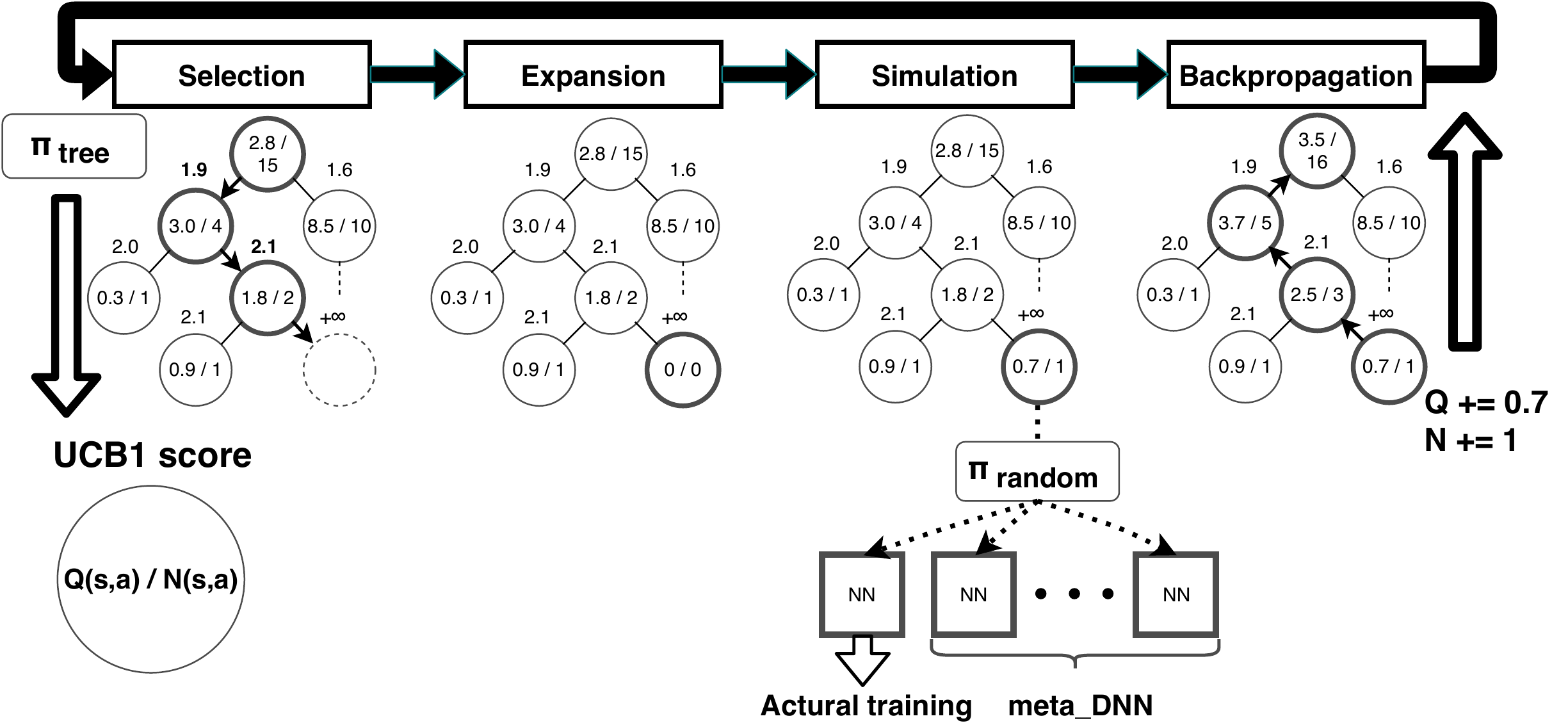}
\caption{\textbf{Overview of AlphaX search procedures:} explanations of four steps are in sec.\ref{sec:search}.}
\label{fig:mcts-procedures}
\end{figure}

\textit{Bayesian Optimization} (BO) is a popular method for the hyper-parameter search \cite{hutter2011sequential,snoek2012practical,thornton2013auto,kandasamy2015high}; It is proven to be an effective black-box optimization technique for small scale problems, e.g. finding good hyper-parameters for Stochastic Gradient Descent (SGD). In a large-scale problem, it demands a good, sophisticated high-dimensional representation kernel to work, requires calculating the inverse of a covariance matrix $\mathcal{O}(n^{2.376})$ that quadratically increases with samples ($n$), so BO is not a practical algorithm for NAS.


\textit{Reinforcement Learning (RL)}: Several RL techniques have been investigated for NAS \cite{Baker2016,Zoph2016}.
Baker et al. proposed a Q-learning agent to design network architectures \cite{Baker2016}. The agent takes a $\epsilon$-greedy policy: with probability $1-\epsilon$, it chooses the action that leads to the best expected return (i.e. accuracy) estimated by the current model, otherwise uniformly chooses an action. Zoph et al. built an RNN agent trained with Policy Gradient to design CNN and LSTM \cite{Zoph2016}. However, directly maximizing the expected reward in vanilla Policy Gradient could lead to local optimal solution~\cite{nachum2016improving}. In comparison, MCTS records both expected return and visitation counts of states to balance between exploration and exploitation at the state level.

\textit{Hill Climbing (HC)}: Elsken et al. proposed a simple hill climbing for NAS \cite{elsken2017simple}. Starting from an architecture, they train every descendent network before moving to the best performing child. Liu et al. deployed a beam search which follows a similar procedure to hill climbing but selects the top-K architectures instead of only the best \cite{liu2017progressive}.
HC is akin to the vanilla Policy Gradient tending to trap into a local optimum from which it can never escape, while MCTS demonstrates provable convergence toward the global optimal given enough time\cite{kocsis2006bandit}.

\textit{Evolutionary Algorithm (EA)}: Evolutionary algorithms represent each neural network as a string of genes and search the architecture space by mutation and recombinations \cite{miller1989designing,Stanley2002,stanley2007compositional,jozefowicz2015empirical,fernando2016convolution,Liu2017a,Real2017,xie2017genetic,miikkulainen2017evolving,Suganuma2017,real2018regularized}.
Strings which represent a neural network with top performance are selected to generate child models. The selection process is in lieu of exploitation, and the mutation is to encourage exploration. Still, GA algorithms do not consider the visiting statistics at individual states, and its performance is on par with \textit{random search} for lacking an online model to inform decisions.

\section{AlphaX: A Scalable MCTS Design Agent}

\begin{figure}[t]
\centering 
\subfloat[][NASNet Cell]{\includegraphics[height=1.1in]{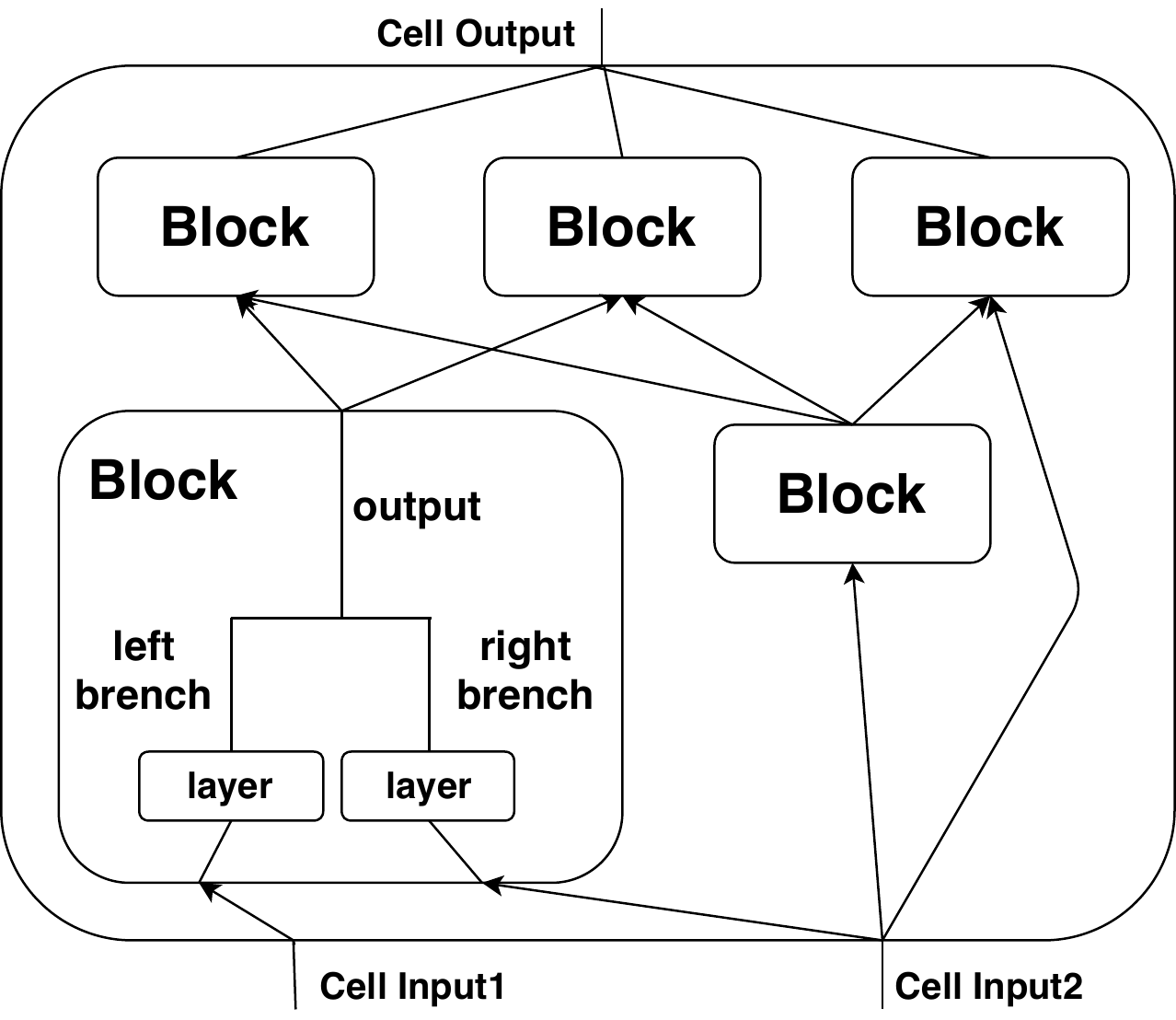}\label{design_domain}} \quad
\subfloat[][NASBench DAG]{\includegraphics[height=1.1in]{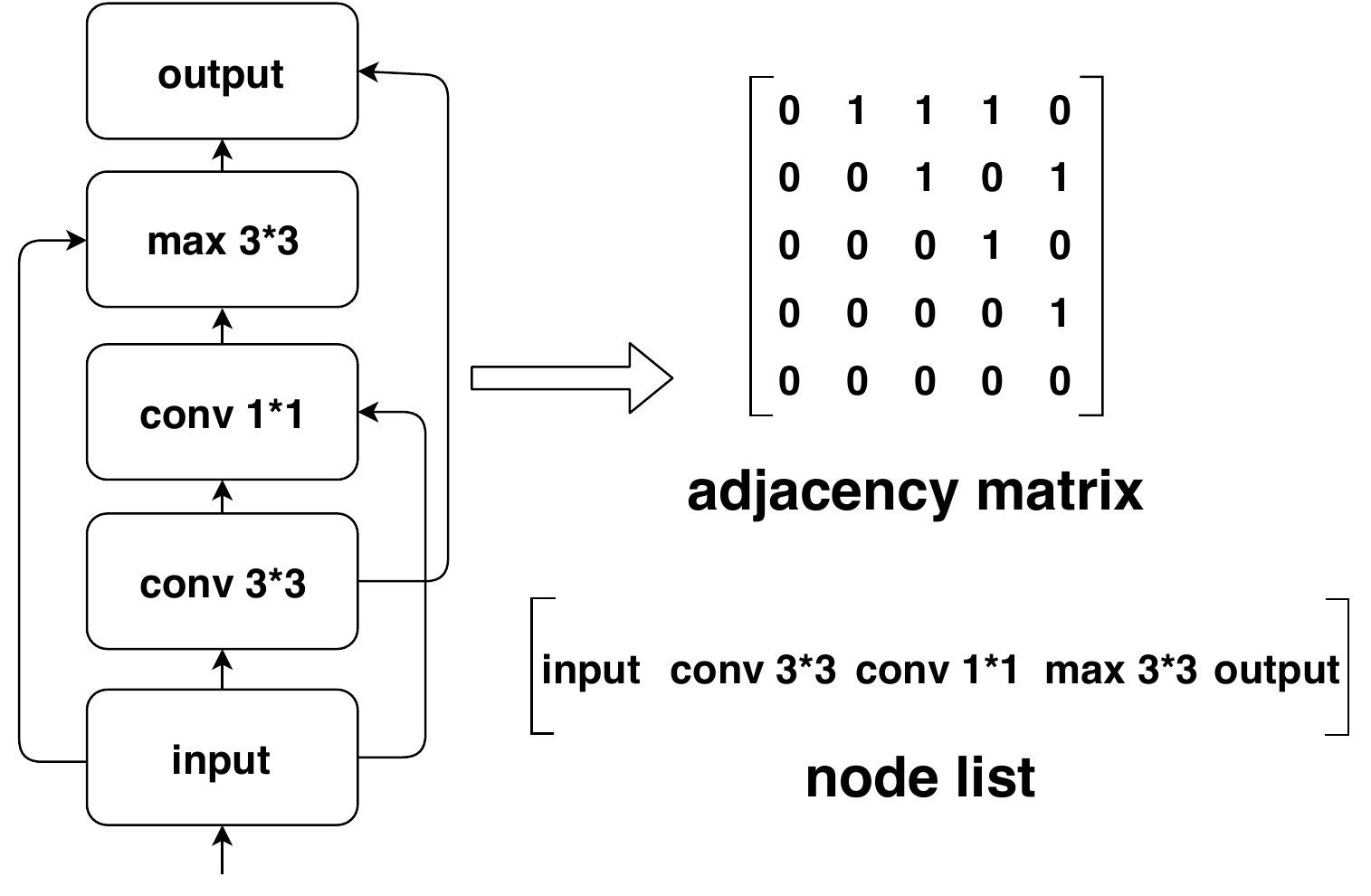}\label{MCTS_terminal_state}}  \quad
\caption{ \textbf{Design space}: (a) the cell structure of NASNet and (b) the DAG structure of NASBench-101. Then the network is constructed by stacking multiple Cells or DAGs. }
\label{nonlinear_connections}
\end{figure}

\subsection{Design, State and Action Space}
\label{content:design_space}
\textbf{Design Space}: the neural architectures for different domain tasks, e.g. the object detection and the image classification, follow fundamentally different designs. This renders different design spaces for the design agent. AlphaX is flexible to support various search spaces with an intuitive state and action abstraction. Here we provide a brief description of two search spaces used in our experiments.  

\begin{itemize}
\item \textit{NASNet Search Space}: \cite{zoph2017learning} proposes searching a hierarchical Cell structure as shown in Fig.\ref{design_domain}. There are two types of Cells, Normal Cell ($NCell$) and Reduction Cell ($RCell$). Normal Cell maintains the input and output dimensions with the padding, while Reduction Cell reduces the height and width by half with the striding. Then, the network is constituted by stacking multiple cells. 

\item \textit{NASBench Search Space}: \cite{ying2019bench} proposes searching a small Direct Acyclic Graph (DAG) with each node representing a layer and each edge representing the inter-layer dependency. Similarly, the network is constituted by stacking multiple such DAGs. 
\end{itemize}

\label{sec:state}

\textbf{State Space}: a state represents a network architecture, and AlphaX utilizes states (or nodes) to keep track of past trails to inform future decisions. We implement a state as a map that defines all the hyper-parameters for each network layers and their dependencies. We also introduce a special terminal state to allow for multiple actions. All the other states can transit to the terminal state by taking the terminal action, and the agent only trains the network, from which it reaches the terminal. With the terminal state, the agent freely modifies the architecture before reaching the terminal. This enables multiple actions for the design agent to bypass shallow architectures.

\textbf{Action Space}: an action morphs the current network architecture, i.e. current state, to transit to the next state. It not only explicitly specifies the inter-layer connectivity, but also all the necessary hyper-parameters for each layer. Unlike games, actions in NAS are dynamically changing w.r.t the current state and design spaces. For example, AlphaX needs to leverage the current DAG (state) in enumerating all the feasible actions of 'adding an edge'. In our experiments, the actions for the NASNet search domain are adding a new layer in the left or right branch of a $Block_i$ in a $Cell$, creating a new $Block$ with different input combinations, and the terminating. The actions for the NASBench search domain are either adding a node or an edge, and the terminating.

\subsection{Search Procedure}

\label{sec:search}

This section elaborates the integration of MCTS and metaDNN in AlphaX. The MCTS is to analyze the most promising move at a state, while the meta-DNN is to learn the sampled architecture performance and to generalize to unexplored architectures so that MCTS can simulate many rollouts with only an actual network training in evaluating a new node. The superior search efficiency of AlphaX is due to balancing the exploration and exploitation at the finest granularity, i.e. state level, by leveraging the visiting statistics. Each node tracks these two statistics: 1) $N(s, a)$ counts the selection of action $a$ at state $s$; 2) $Q(s, a)$ is the expected reward after taking action $a$ at state $s$, and intuitively $Q(s, a)$ is an estimate of how promising this search direction. Fig.\ref{fig:mcts-procedures} demonstrates a typical searching iteration in AlphaX, which consists of \textit{Selection}, \textit{Expansion}, \textit{Meta-DNN assisted Simulation}, and \textit{Backpropagation}. We elucidate each step as follows.


\textbf{\textit{Selection}} traverses down the search tree to trace the current most promising search path. It starts from the root and stops till reaching a leaf. At a node, the agent selects actions based on UCB1 \cite{auer2002finite}:
\begin{equation}
	\label{ucb1_update}
	\pi_{tree}(s) = \text{arg}\max_{a \in A} \left( \frac{Q(s, a)}{N(s, a)} + 2 c \sqrt{\frac{2 \log N(s)}{N(s, a)}} \right),
\end{equation}
where $N(s)$ is the number of visits to the state $s$ (i.e. $N(s) = \sum_{a \in A} N(s, a)$), and $c$ is a constant. The first term ($\frac{Q(s, a)}{N(s, a)} $) is the exploitation term estimating the expected accuracy of its descendants. The second term ($2 c \sqrt{\frac{2 \log N(s)}{N(s, a)}}$) is the exploration term encouraging less visited nodes. The exploration term dominates $\pi_{tree}(s)$ if $N(s, a)$ is small, and the exploitation term otherwise. As a result, the agent favors the exploration in the beginning until building proper confidences to exploit. $c$ controls the weight of exploration, and it is empirically set to 0.5. We iterate the tree policy to reach a new node.


\textbf{\textit{Expansion}} adds a new node into the tree. $Q(s, a)$ and $N(s, a)$ are initialized to zeros. $Q(s, a)$ will be updated in the simulation step.

\textbf{\textit{Meta-DNN assisted Simulation}} randomly samples the descendants of a new node to approximate $Q(s, a)$ of the node with their accuracies.
The process is to estimate how promising the search direction rendered by the new node and its descendants. The simulation starts at the new node. The agent traverses down the tree by taking the uniform-random action until reaching a terminal state, then it dispatches the architecture for training.

The more simulation we roll, the more accurate estimate of this search direction we get. However, we cannot conduct many simulations as network training is extremely time-consuming.
AlphaX adopts a novel hybrid strategy to solve this issue by incorporating a meta-DNN to predict the network accuracy in addition to the actual training. We delay the introduction of meta-DNN to sec.\ref{sec:metadnn}.
Specifically, we estimate $q = Q(s, a)$ with
\begin{equation}
	\label{reward}
	Q(s, a) \leftarrow \left( Acc(sim_0(s')) + \frac{1}{k} \sum_{i=1..k} Pred(sim_i(s')) \right) / 2
\end{equation}
where $s' = s + a$, and $sim(s')$ represents a simulation starting from state $s'$. $Acc$ is the actually trained accuracy in the first simulation, and $Pred$ is the predicted accuracy from Meta-DNN in subsequent $k$ simulations. If a search branch renders architectures similar to previously trained good ones, Meta-DNN updates the exploitation term in Eq.\ref{ucb1_update} to increase the likelihood of going this branch.


\textbf{\textit{Backpropagation}} back-tracks the search path from the new node to the root to update visiting statistics. Please note we discuss the sequential case here, and the backpropagation will be split into two parts in the distributed setting (sec.\ref{sec:distributed}). 
With the estimated $q$ for the new node, we iteratively back-propagate the information to its ancestral as:
\begin{equation}
	\begin{split}
	& Q(s, a) \leftarrow Q(s, a) + q, \quad
	 N(s, a) \leftarrow N(s, a) + 1 \quad \\
	& s \leftarrow parent(s), \quad
	 a \leftarrow \pi_{tree}(s)
	\end{split}
\end{equation}
until it reaches the root node.

\begin{figure}
\vspace{-0.2in}
\centering 
\includegraphics[height=1.7in]{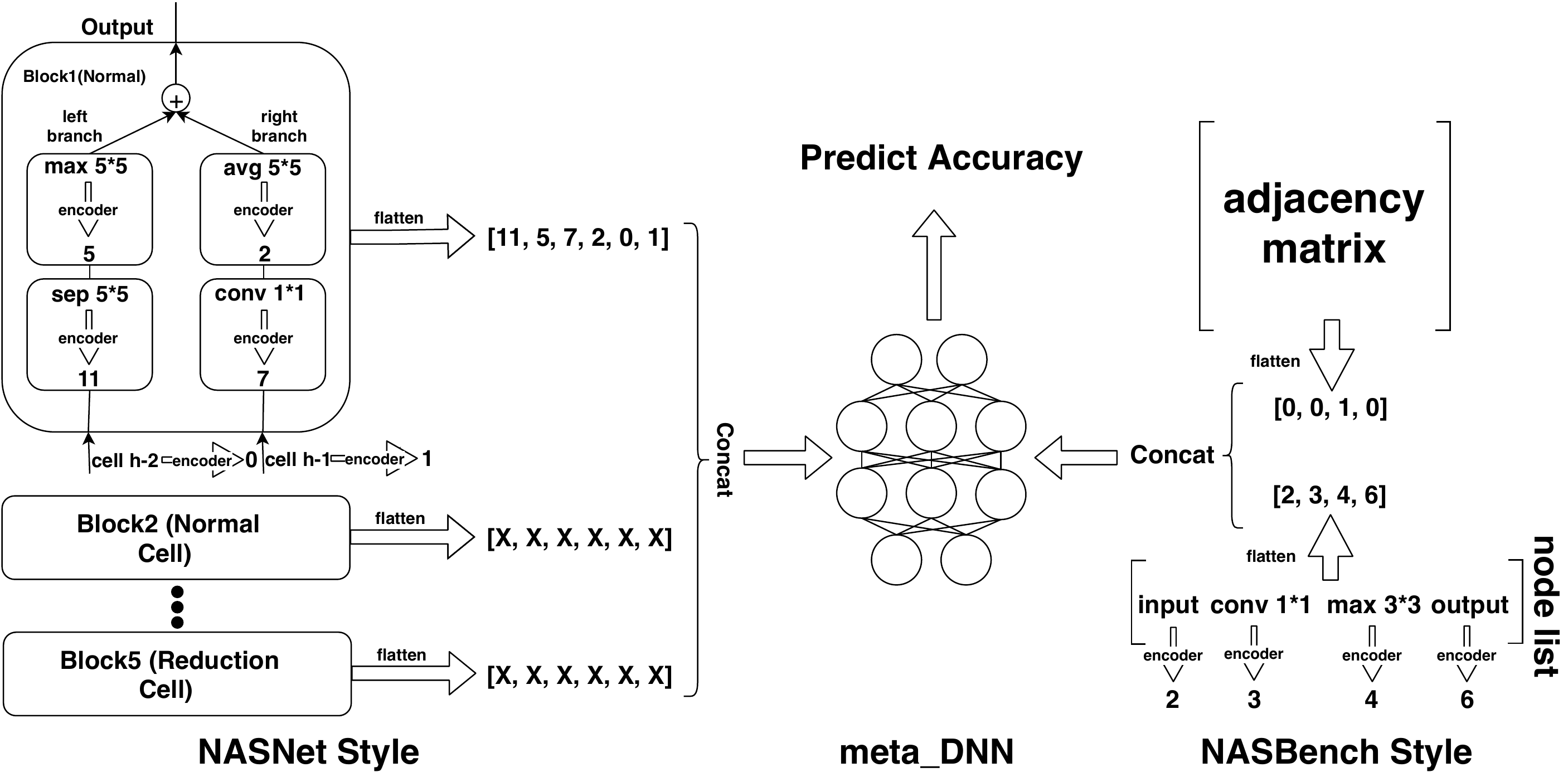}
\caption{Encoding scheme of NASBench and NASNet.}
\label{fig:network-encoding}
\end{figure}

\subsection{The design of Meta-DNN and its related issues}
\label{sec:metadnn}

Meta-DNN intends to generalize the performance of unseen architectures based on previously sampled networks. It provides a practical solution to accurately estimate a search branch with many simulations without involving the actual training (see the metaDNN assisted simulation for details). New training data is generated as AlphaX advances in the search. So, the learning of Meta-DNN is end-to-end. The input of Meta-DNN is a vector representation of architecture, while the output is the prediction of architecture performance, i.e. test accuracy. 

The coding scheme for NASNet architectures is as follows: we use 6-digits vector to code a $Block$; the first two digits represent up to two layers in the left branch, and the 3rd and 4th digits for the right branch. Each layer is represented by a number in $[1, 12]$ to represent 12 different layers, and the specific layer code is available in Appendix TABLE.\ref{app:layers_used}. We use 0 to pad the vector if a layer is absent. The last two digits represent the input for the left and right branch, respectively. For the coding of block inputs, 0 corresponds to the output of the previous $Cell$, 1 is the previous, previous $Cell$, and $i+2$ is the output of $Block_i$. If a block is absent, it is [0,0,0,0,0,0]. The left part of Fig.\ref{fig:network-encoding} demonstrates an example of NASNet encoding scheme. A $Cell$ has up to 5 blocks, so a vector of 60 digits is sufficient to represent a state that fully specifies both $RCell$ and $NCell$. The coding scheme for NASBench architectures is a vector of flat adjacency matrix, plus the nodelist. Similarly, we pad 0 if a layer or an edge is absent. The right part of Fig.\ref{fig:network-encoding} demonstrates an example of NASBench encoding scheme. Since NASBench limits nodes $\leq$ 7, 7$\times$7 (adjacency matrix)+ 7 (nodelist) = 56 digits can fully specify a NASBench architecture.

Now we cast the prediction of architecture performance as a regression problem. Finding a good metaDNN is heuristically oriented and it should vary from tasks to tasks. We calculate the correlation between predicted accuracies and true accuracies from the sampled architectures in evaluating the design of metaDNN. Ideally, the metaDNN is expected to rank an unseen architecture in roughly similar to its true test accuracy, i.e. corr = 1. Various ML models, such as Gaussian Process, Neural Networks, or Decision Tree, are candidates for this regression task. We choose Neural Networks as the backbone model for its powerful generalization on the high-dimensional data and the online training capability. More ablations studies for the specific choices of metaDNN is available in sec.\ref{sec:metadnn_design}.

\subsection{ Transfer Learning }
As MCTS incrementally builds a network with primitive actions, networks falling on the same search path render similar structures. This motivates us to incorporate Transfer Learning in AlphaX to speedup network evaluations. In simulation (Fig.~\ref{fig:mcts-procedures}), AlphaX recursively traverses up the tree to find a previously trained network with the minimal edit distance to the newly sampled network. Then we transfer the weights of overlapping layers, and randomly initialize new layers. In the pre-training, we train every samples for 70 epochs, while we train an architecture for 20 epochs if the parent weights are available. We provide an ablation study in Fig.~\ref{fig:transfer_learning} to justify the design.    



\subsection{Distributed AlphaX}
\label{sec:distributed}

\begin{figure}
  \begin{center}
    \includegraphics[width=0.7\columnwidth]{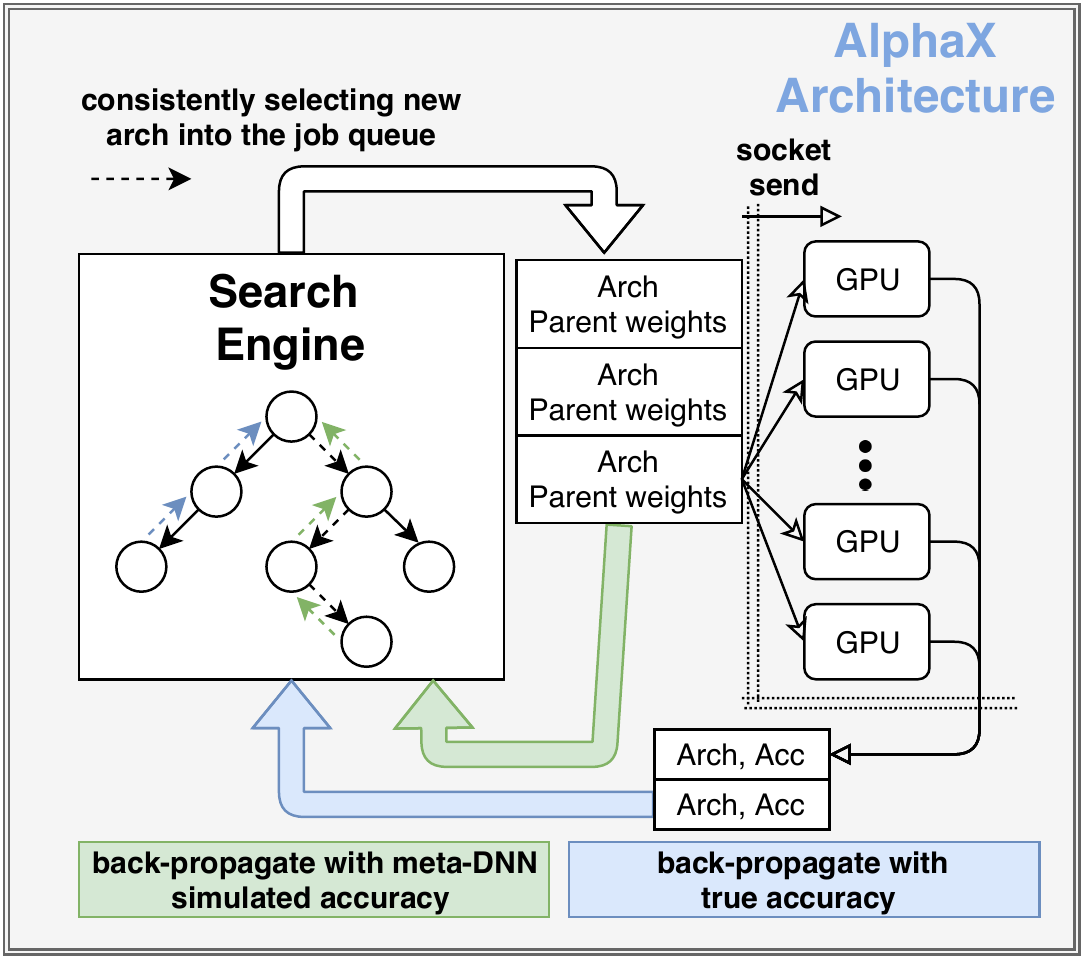}
  \end{center}
  \caption{\textbf{Distributed AlphaX}: we decouple the original back-propagation into two parts: one uses predicted accuracy (green arrow), while the other uses the true accuracy (blue arrow).  The pseudocode for the whole system is available in Appendix Sec.\ref{ap:pseudocode} }
  \vspace{-0.1in}
  \label{distributed_alphaX}
\end{figure}

It is imperative to parallelize AlphaX to work on a large scale distributed systems to tackle the computation challenges rendered by NAS.
Fig.\ref{distributed_alphaX} demonstrates the distributed AlphaX. There is a master node exclusively for scheduling the search, while there are multiple clients (GPU) exclusively for training networks. The general procedures on the server side are as follows: 1) The agent follows the selection and expansion steps described in Fig.\ref{fig:mcts-procedures}. 2) The simulation in MCTS picks a network $arch_n$ for the actual training, and the agent traverses back to find the weights of parent architecture having the minimal edit distance to $arch_n$ for transfer learning; then we push both $arch_n$ and parent weights into a job queue. We define $arch_n$ as the selected network architecture at iteration $n$, and $rollout_{-}from(arch_n)$ as the node which it started the rollout from to reach $arch_n$. 3) The agent {\it preemptively backpropagates} $\hat{q} \leftarrow \frac{1}{k} \sum_{i=1..k} Pred(sim_i(s'))$ based only on predicted accuracies from the Meta-DNN at iteration $n$.
\begin{equation}
	\begin{split}
	&Q(s, a) \leftarrow Q(s, a) + \hat{q}, \quad
	N(s, a) \leftarrow N(s, a) + 1, \quad \\
	&s \leftarrow parent(s), \quad
	a \leftarrow \pi_{tree}(s).
	\end{split}
\end{equation}
4) The server checks the receive buffer to retrieve a finished job from clients that includes $arch_z$, $acc_z$.
Then the agent starts the second backpropagation to propagate $q \leftarrow \frac{acc_z + \hat{q}}{2}$ (Eq. \ref{reward}) from the node the rollout started ($s \leftarrow rollout_{-}from(arch_z)$) to replace the backpropagated $\hat{q}$ with $q$:

\begin{equation}
	\begin{split}
	Q(s, a) \leftarrow Q(s, a) + q - \hat{q}, \quad \\
	s \leftarrow parent(s), \quad 
	a \leftarrow \pi_{tree}(s).
	\end{split}
\end{equation}

The client constantly tries to retrieve a job from the master job queue if it is free. It starts training once it gets the job, then it transmits the finished job back to the server. So, each client is a dedicated trainer. We also consider the fault-tolerance by taking a snapshot of the server's states every few iterations, and AlphaX can resume the searching from the breakpoint using the latest snapshot.

\begin{figure}
  \begin{center}
    \includegraphics[width=0.49\columnwidth]{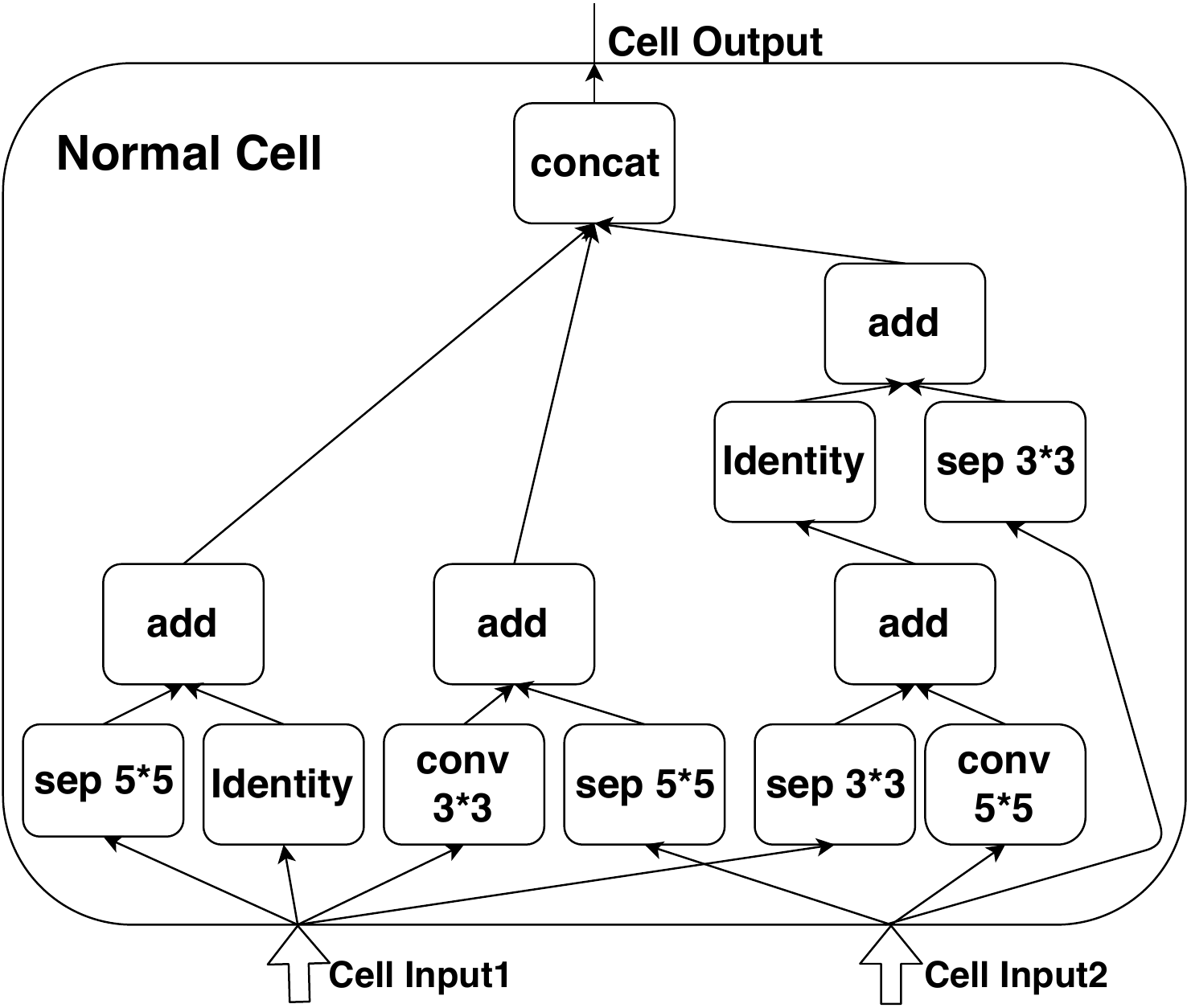}
    \includegraphics[width=0.49\columnwidth]{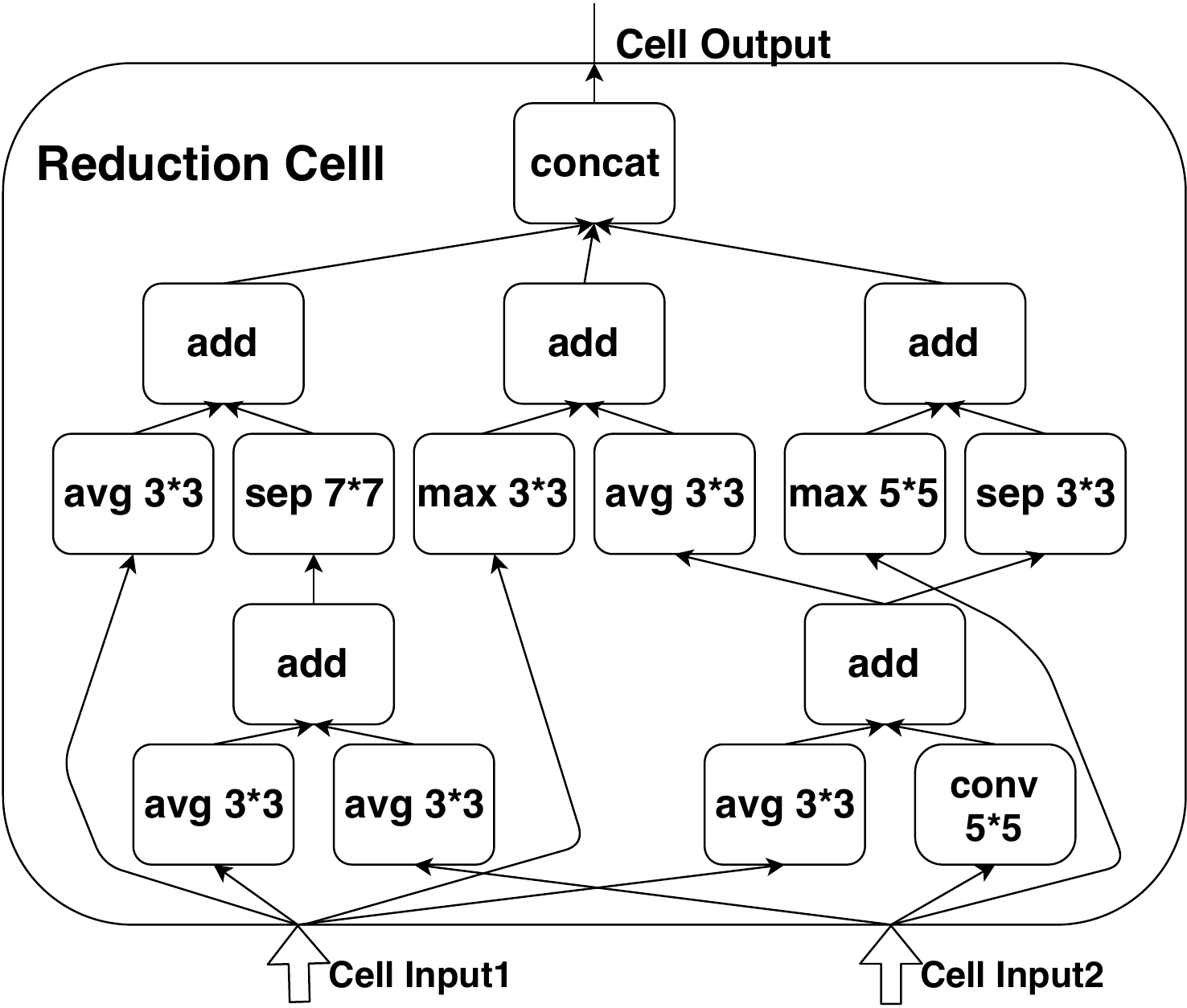}
  \end{center}
  \caption{the $RCell$ and $NCell$ that yield the highest accuracy in the search.}
  \label{best-arch}
\end{figure}

\section{Experiment}

\subsection{Evaluations of Architecture Search}
\textit{\textbf{Experiment setup}}:  An anonymized implementation of AlphaX to search on NASBench-101 for the reviewers are at \cite{alphax-implementation}. Appendix Sec.\ref{ap:experiment_setup} provides the details of experiment setup.

  \begin{figure}[t]
    \begin{center}
      \subfloat[][best acc progression]{ \includegraphics[height=3.3cm]{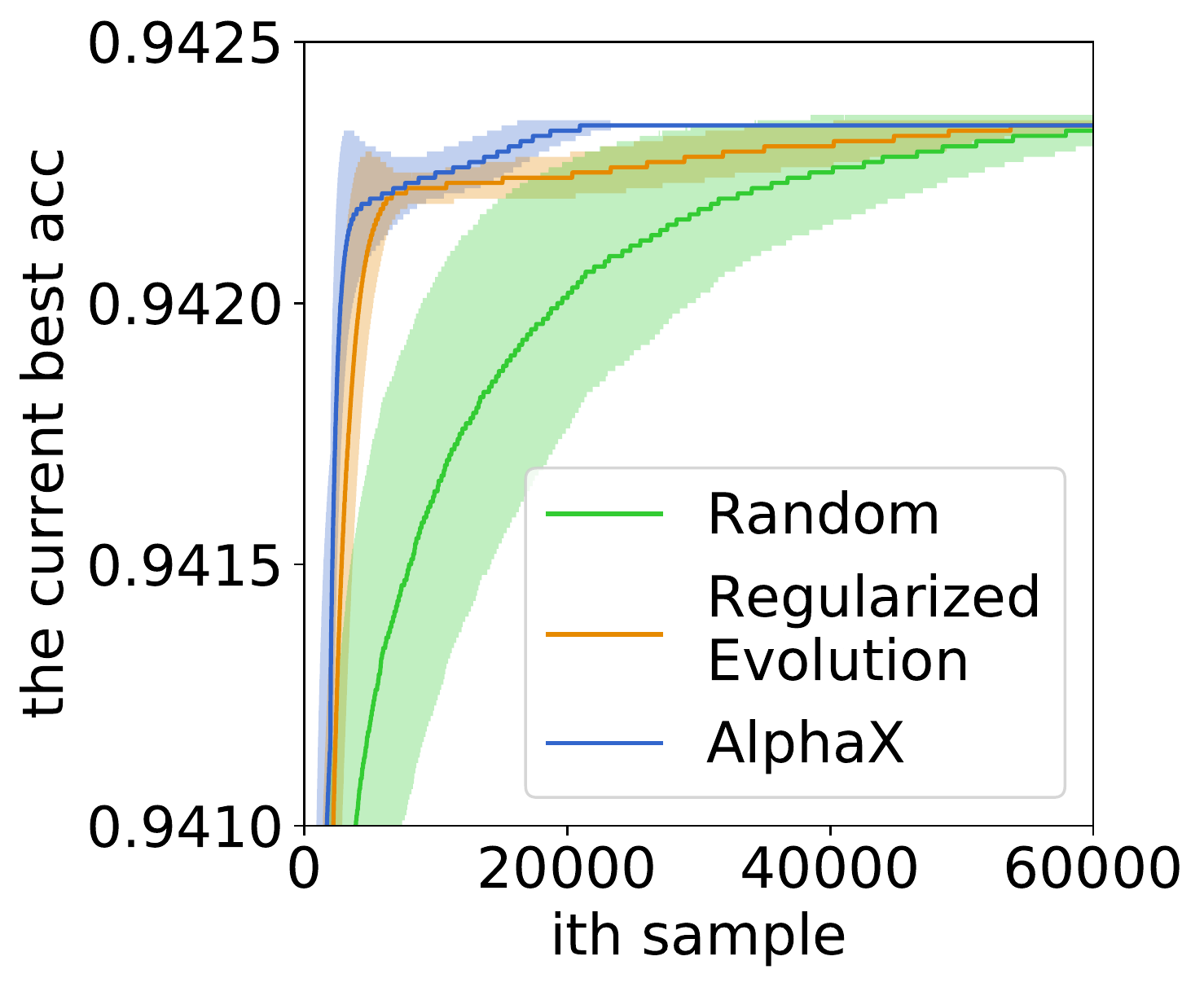}\label{fig:nasbench_best_trace}}   \quad
      \subfloat[][\#samples to the best]{ \includegraphics[height=3.3cm]{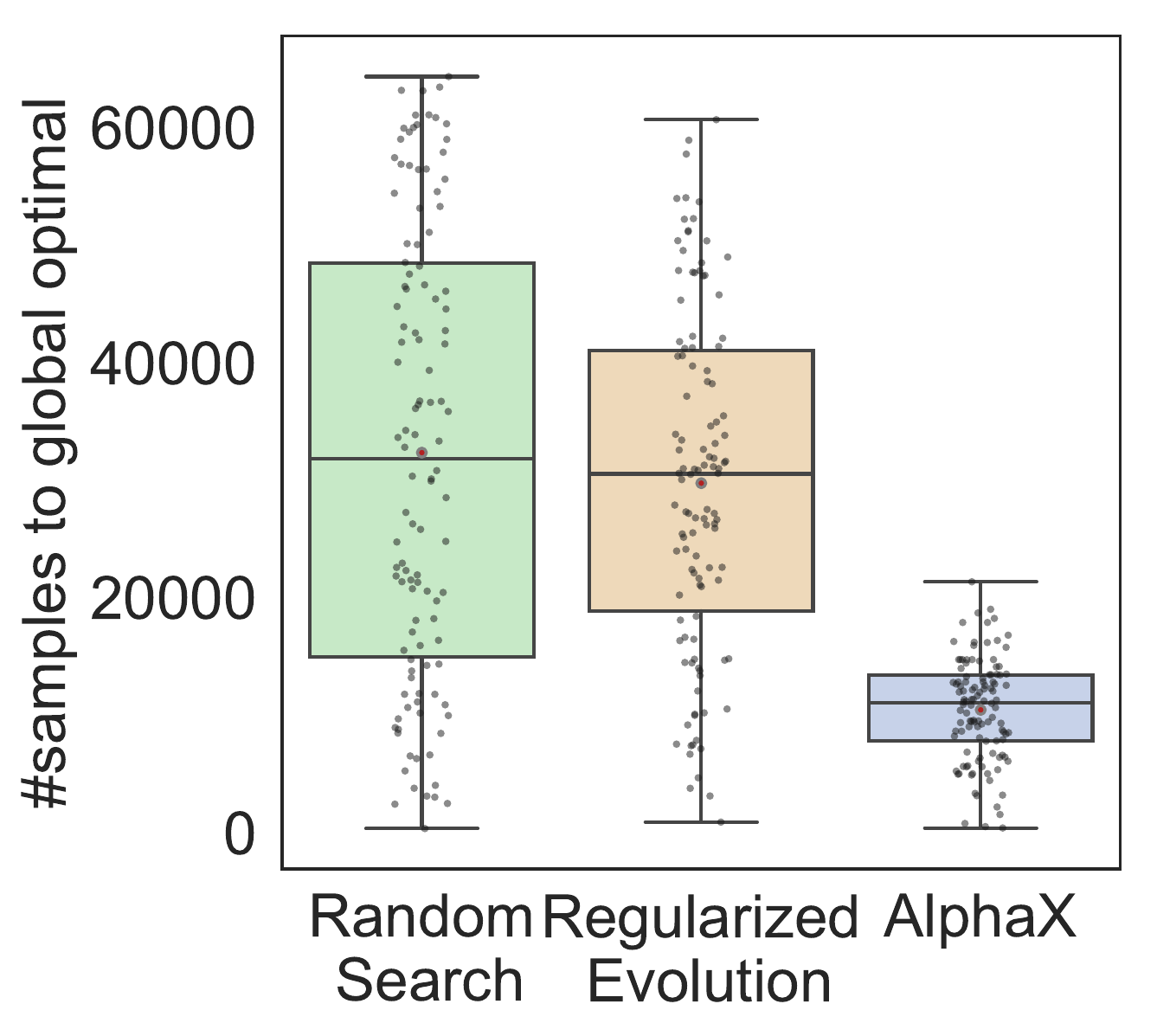}\label{fig:nasbench_samples_to_best}}
    \end{center}
    \caption{\textbf{Finding the global optimum on NASBench-101}: AlphaX is 3x, 2.8x faster than Random Search and Regularized Evolution on NASBench-101 (nodes $\leq 6$). The results are from 200 trails with different random seeds. }
    \label{fig:nasbench-results}
  \end{figure}

\subsubsection{ Finding the Global Optimal on NASBench-101 }
Searching NASNet requires training thousands of networks, so it is not computationally feasible to conduct many trials to fairly evaluate a search algorithm. Current literatures mainly evaluate NAS algorithms with the final test accuracy, while \cite{sciuto2019evaluating} has shown many state-of-the-art NAS algorithms, e.g. DARTS \cite{liu2018darts}, NAO\cite{luo2018neural}, ENAS \cite{pham2018efficient}, cannot even outperform Random Search in the same setting. To truly evaluate a search algorithm, and to bypass the computational challenges, Christ et al collected NASBench \cite{ying2019bench} that enumerates all the possible DAGs of nodes $\leq7$, constituting of (420k+) networks and their final test accuracies.

In our experiments, we limit the maximal nodes in a DAG $\leq$ 6 for repeating each algorithm for 200 trails, and the rest follows the same NASBench setting, i.e. taking a subset of NASBench-101 with 64521 valid networks. The search target is the network with the highest mean test accuracy (global optimal) at 108th epochs, which can be known ahead by querying the dataset. Our evaluation metric is the number of samples to reach the best performance architecture in testing. We choose Random Search (RS) \cite{sciuto2019evaluating} and Regularized Evolution (RE) \cite{real2018regularized} as the baseline, and experimental results are in 
Fig.\ref{fig:nasbench-results}. We have run each algorithm for 200 independent trials, and each trail is a new search fed with a different random seed. The search terminates once it reaches the target (the global optimal). Fig.\ref{fig:nasbench-results} demonstrates AlphaX is 2.8x and 3x faster than RS and RE, respectively. As we analyzed in Fig.\ref{fig:teaser}, Random Search lacks an online model. Regularized Evolution only mutates on top-k performing models, while MCTS explicitly builds a search tree to dynamically trade off the exploration and exploitation at individual states. Please note that the slight difference in Fig.\ref{fig:nasbench_best_trace} actually reflects a huge gap in speed as indicated by Fig.\ref{fig:nasbench_samples_to_best}. This is due to the minor difference (within 0.5\%) between the near optimal architectures and the global optimal.

\subsubsection{ Open domain search \cite{zoph2017learning}}
we perform the search on CIFAR-10 using 10 NVIDIA 1080 TI. One GPU works as a server, while the rest work as clients. The client-server communications are through python sockets. To sample more architectures within limited resources, we early terminate the training at the 70th on each search clients. Then we rank architectures by their preliminary accuracies to perform an additional 530 epochs on good candidates. In acquiring the final accuracy, cutout is applied \cite{cosine_restart} using 1 random crop of size $16\times16$. For the subsequent ImageNet training, we construct the network for ImageNet with searched $RCell$ and $NCell$ according to Fig.2 in \cite{zoph2017learning}. We set up the ImageNet training using the standard mobile configuration with the input image size of ($224\times224$)\cite{zoph2017learning}. More details are available in the appendix. AlphaX sampled 1000 networks, and we selected the top 20 networks in the pre-training to fine-tune another 530 epochs.  Fig.\ref{best-arch} demonstrates the architecture that yields the highest accuracy after fine-tuning.

\begin{table}[t]
\setlength{\tabcolsep}{0.4em}
  \scriptsize
  \label{layers_used}
  \centering
  \begin{tabular}{l l l l l l}
    \toprule
         \textbf{Model}      & \textbf{Space}  & \textbf{Params}  & \textbf{Err} & \textbf{GPU days} & \textbf{M}\\
    \midrule
		  NASNet-A+cutout    \cite{zoph2017learning}        & NASNet    & 3.3M   & 2.65 & 2000 & 20000 \\
		  AmoebaNet-B+cutout \cite{real2018regularized}   & NASNet    & 2.8M   & $2.50_{\pm0.05}$ & 3150 & 27000 \\
  		  DARTS+cutout       \cite{liu2018darts}                & NASNet    & 3.3M   & $2.76_{\pm0.09}$  & 4  & -   \\
		  RENASNet+cutout    \cite{chen2019renas} & NASNet & 3.5M & $2.88_{\pm0.02}$ & 6 & 4500 \\
		  \midrule
		  AlphaX+cutout (32 filters)  		& NASNet			  & 2.83M & $2.54_{\pm0.06}$ & 15  & 1000   \\
		  \midrule
		  PNAS               \cite{liu2017progressive}           & NASNet	  & 3.2M   & $3.41_{\pm0.09}$  & 225  & 1160   \\
		  ENAS               \cite{pham2018efficient}           & NASNet    & 4.6M   & 3.54  & 0.45 &  - \\
		  NAONet             \cite{luo2018neural} & NASNet & 10.6M & 3.18 & 200 & 1000          \\
		  \midrule
          AlphaX (32 filters)  		& NASNet			  & 2.83M & $3.04_{\pm0.03}$         & 15  & 1000   \\
		  
	\midrule
  		  NAS v3\cite{Zoph2016}				     & NAS        	 & 7.1M   & 4.47 & 22400 & 12800  \\ 
          Hier-EA	  \cite{liu2017hierarchical} & Hier-EA       & 15.7M  & $3.75_{\pm0.12}$ & 300  & 7000  \\
		  Proxyless-G \cite{cai2018proxylessnas} & ConvNet       & 5.7M.  & 2.08 & 8.33 & N/A \\
	\midrule
		  AlphaX+cutout (64 filters)  		& NASNet			  & 9.36M & $2.06_{\pm0.04}$ & 15  & 1000   \\		  
    \bottomrule
    \label{acc-comps-cifar}
  \end{tabular}
  \caption{The comparisons of our NASNet search results to other state-of-the-art results on CIFAR-10. M is the number of sampled architectures in the search. The cell structure of AlphaX is in Fig.~\ref{best-arch}.}
  
  \footnotetext{* represents cutout}
\end{table}

\begin{table}[t]
\setlength{\tabcolsep}{0.2em}

  \scriptsize
  \label{layers_used}
  \centering
  \begin{tabular}{ l l l l l l}
    \toprule
          \textbf{model} 							& \textbf{multi-adds}      & \textbf{params}& \textbf{latency} & \textbf{top1/top5 err}  \\
    \midrule
		  NASNet-A      \cite{zoph2017learning}     &    564M				   & 5.3M  &  39.68ms & 26.0/8.4 \\
		  AmoebaNet-B   \cite{real2018regularized}  &    555M                  & 5.3M  &  32.15ms & 26.0/8.5 \\
		  DARTS         \cite{liu2018darts}         &    574M                  & 4.7M  &  39.32ms & 26.7/8.7 \\
		  RENASNet      \cite{chen2019renas}        &    574M                  & 4.7M  &  41.21ms & 24.3/7.4 \\
		  PNAS          \cite{liu2017progressive}   &    588M                  & 5.1M  &  40.34ms    & 25.8/8.1 \\
		  Proxyless-G   \cite{cai2018proxylessnas}  &    -                     & -     &  83ms    & 25.4/7.8 \\
	\midrule
		  AlphaX-1  						        &   579M  & 5.4M & 38.56ms & 24.5/7.8   \\
    \bottomrule
    \label{acc-comps-imagenet}
  \end{tabular}
  \caption{The error rate (\%) comparisons of our best-performing architecture to other state-of-the-art results on ImageNet. The network setup follows the mobile setting defined in \cite{zoph2017learning}. }
  
  \vspace{-0.2in}
\end{table}

\begin{figure}
  \begin{center}
    \includegraphics[width=0.8\columnwidth ]{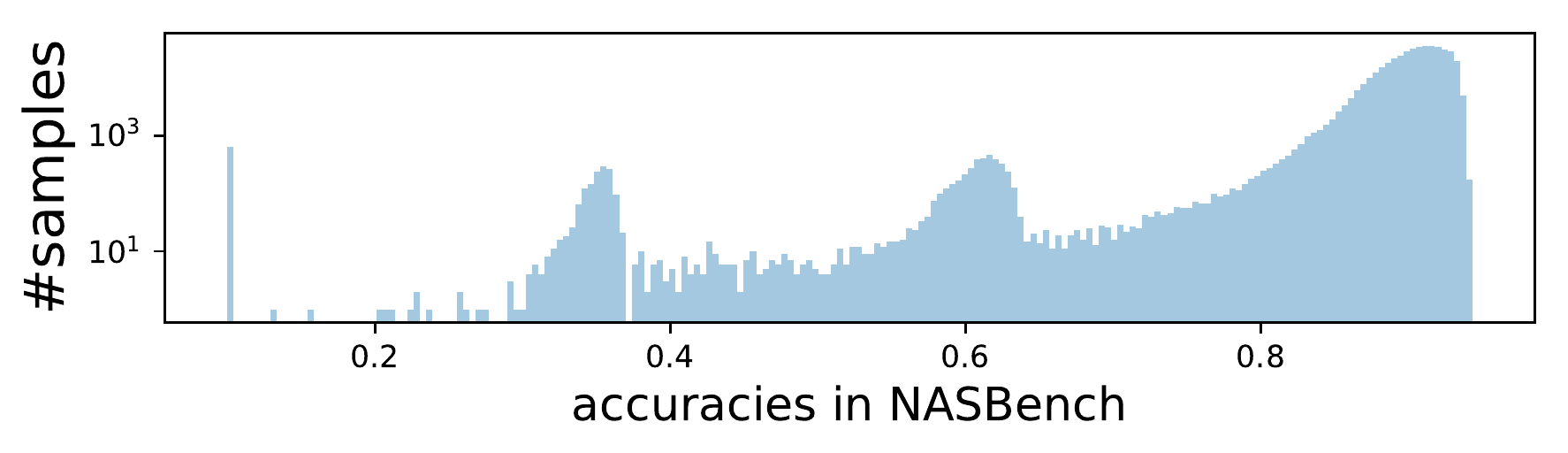}
  \end{center}
  \vspace{-0.3in}
  \caption{Accuracy distribution of networks in NASBench. }
  \label{fig:nasbench_acc_dist}
\end{figure}

\textit{\textbf{Comparisons to State-of-the-Art Results}}: Table.~2 and Table.~3 summarize state-of-the-art results on CIFAR10 and ImageNet, and AlphaX consistently exceeds state-of-the-art methods in both samples (M) and accuracies. Our end-to-end search cost, i.e. GPU days, is on par with SToA methods thanks to the early terminating and transfer learning. Notably, AlphaX achieves the similar accuracy to AmoebaNet with 27x less samples for the case with cutout and filters = 32. Without cutout and filters = 32, AlphaX outperforms NAONet by 0.14\% in the test error with 13.3x less GPU days. Proxyless-G used a different search space, and our result is on par with it after increasing the filters to 64.


  \begin{figure}[t]
  \begin{center}
   \includegraphics[width=0.75\columnwidth, trim={1cm 0cm 1cm 0cm}]{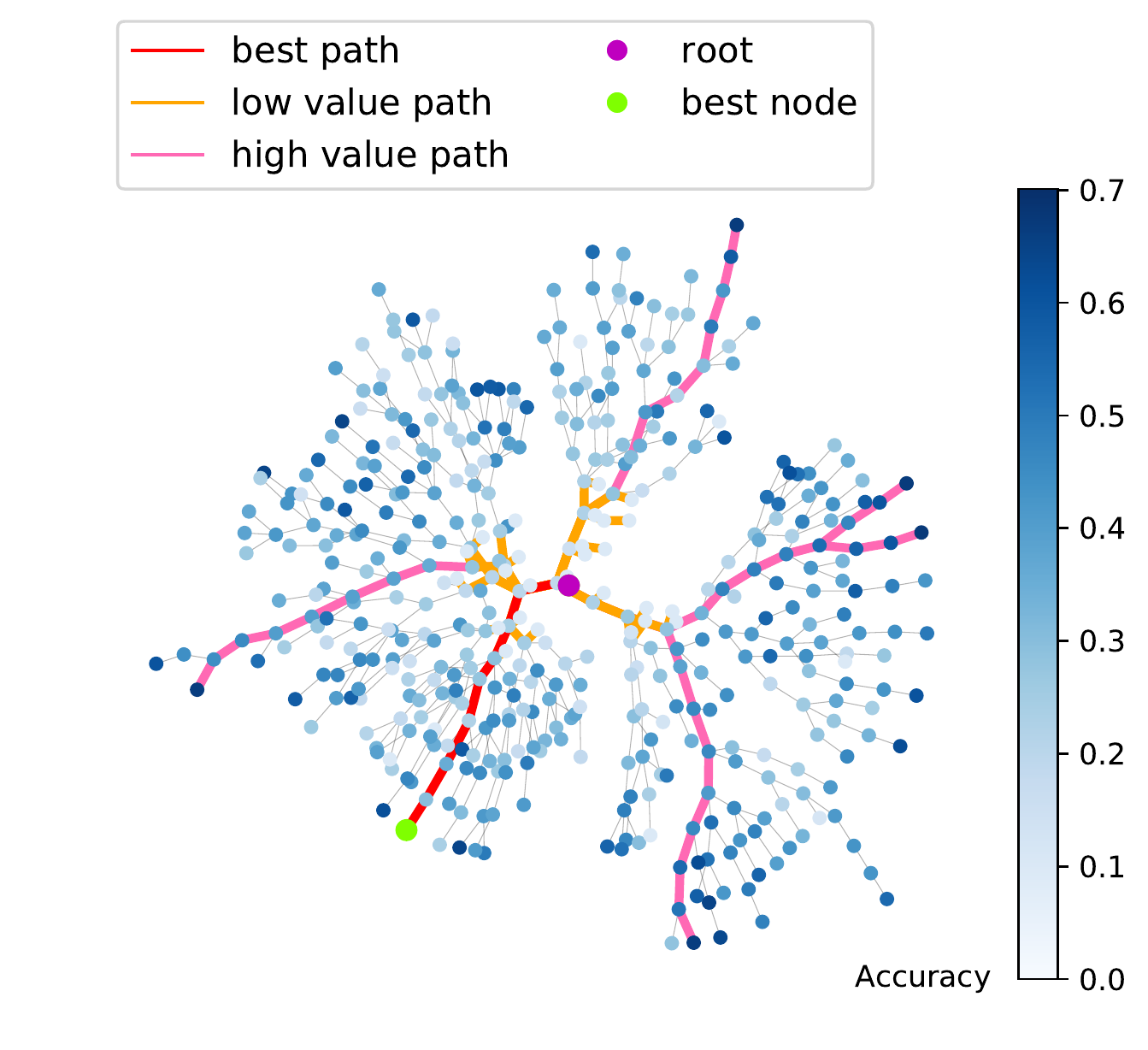}
  \end{center}
  \caption{\textbf{AlphaX search visualization}:each nodes represents a MCTS state; the node color reflects its value, i.e. accuracy, indicating how promising a search branch.  }
  \vspace{-0.1in}
  \label{fig:mcts_viz}
  \end{figure}
  
  \begin{figure}[t]
    \begin{center}
      \subfloat[][best acc progression]{ \includegraphics[height=3.3cm]{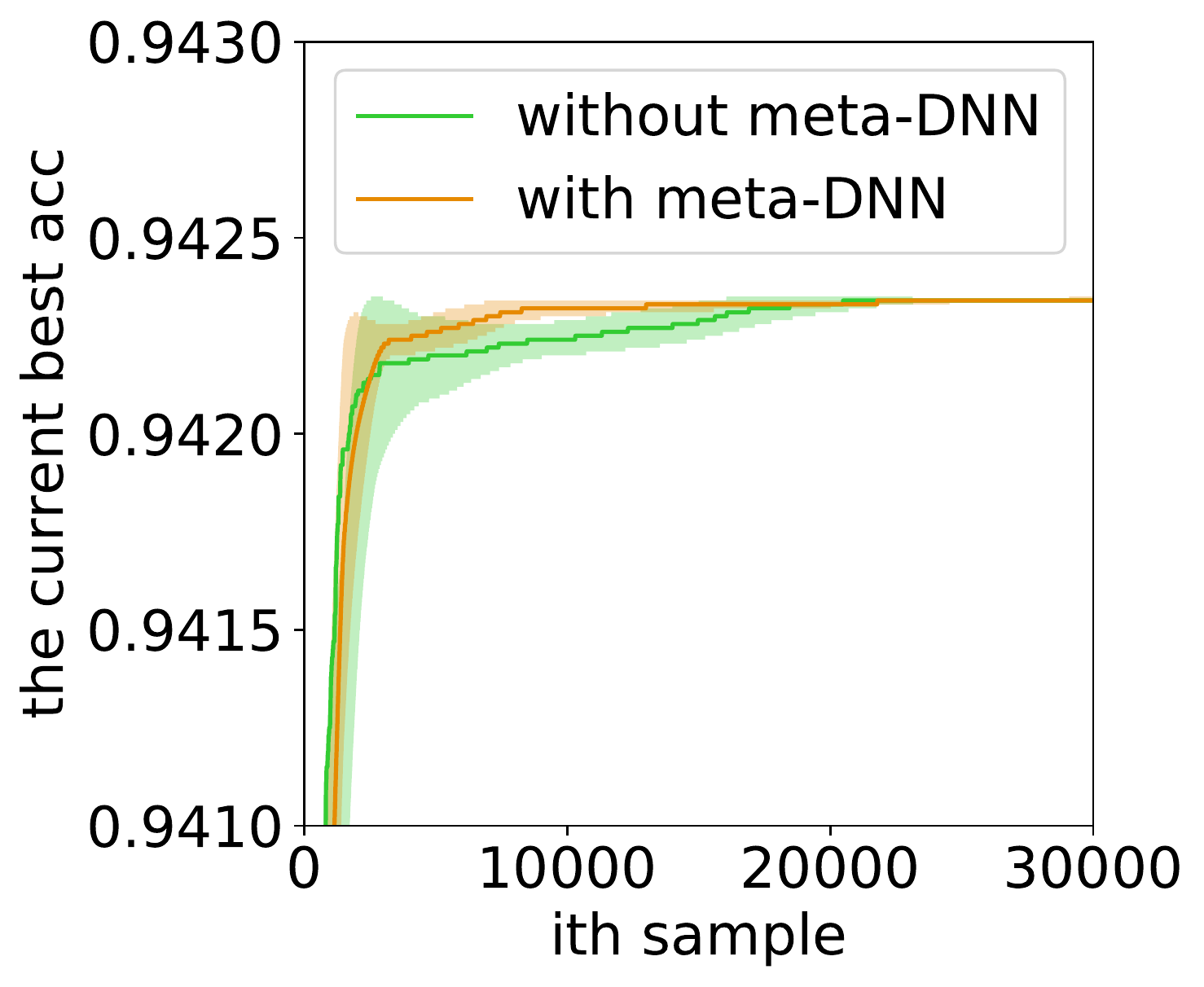}\label{fig:nasbench-metadnn-best-trace}}   \quad
      \subfloat[][\#samples to the best]{ \includegraphics[height=3.3cm]{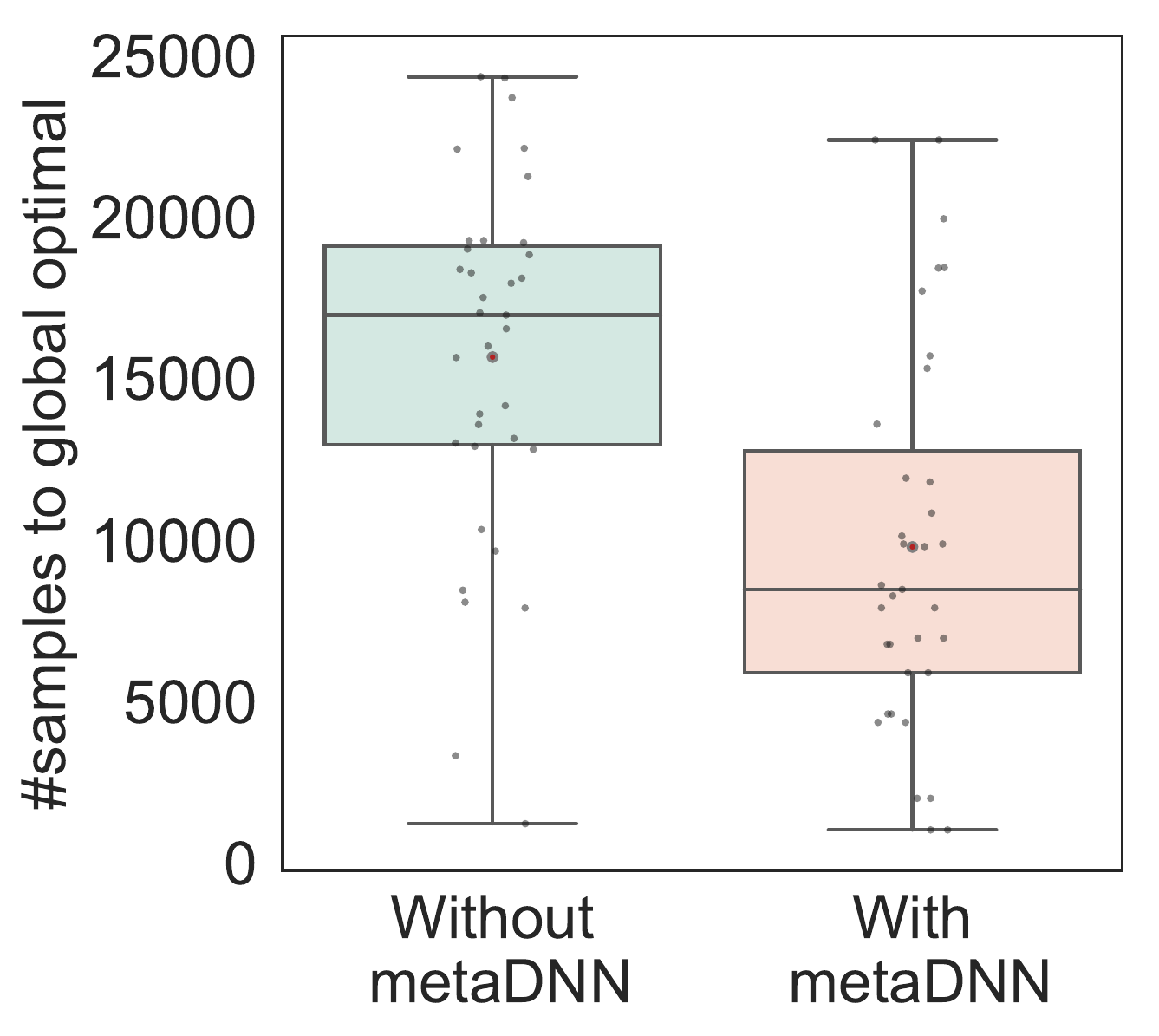}\label{fig:nasbench-metadnn-samples-to-best}}
    \end{center}
    \caption{\textbf{Meta-DNN accelerates the search}: the performance of AlphaX in cases of with/without meta-DNN on NASBench-101. The data is collected from 40 trails.}
    \label{fig:nasbench-metadnn}
  \end{figure}

\subsubsection{ Qualitative Evaluations of AlphaX }
Several interesting insights are observable in Fig.\ref{fig:mcts_viz}.
1) MCTS explicitly builds a search tree to better inform future decisions based on prior rollouts. The choice of actions in MCTS is fine-grained, leveraging both the visiting statistics and value at individual states; while the random or evolutionary search is stateless, utilizing a coarse estimation of search direction (e.g. mutation). Therefore, MCTS is a lot faster than RS and RE in Fig.\ref{fig:nasbench-results}.
2) MCTS invests more on the promising directions (high value path), and less otherwise (low value path). Unlike greedy based algorithms, e.g. hill climbing, MCTS consistently explores the search space as the search tree is similar to a balanced tree in Fig.\ref{fig:mcts_viz}. All of these manifest that MCTS automatically balances the exploration and exploitation in the progress of the search. 3) the best performing network is not necessarily located on the most promising branch, and there are many local optimal branches (dark blue nodes). MCTS is theoretically guaranteed to find the global optimal in this non-convex search space, while PNAS or DARTs can easily trap into a local optimal.

  \begin{figure}[t]
  \centering
  \subfloat[][RNN train]{ \includegraphics[width=0.32\columnwidth]{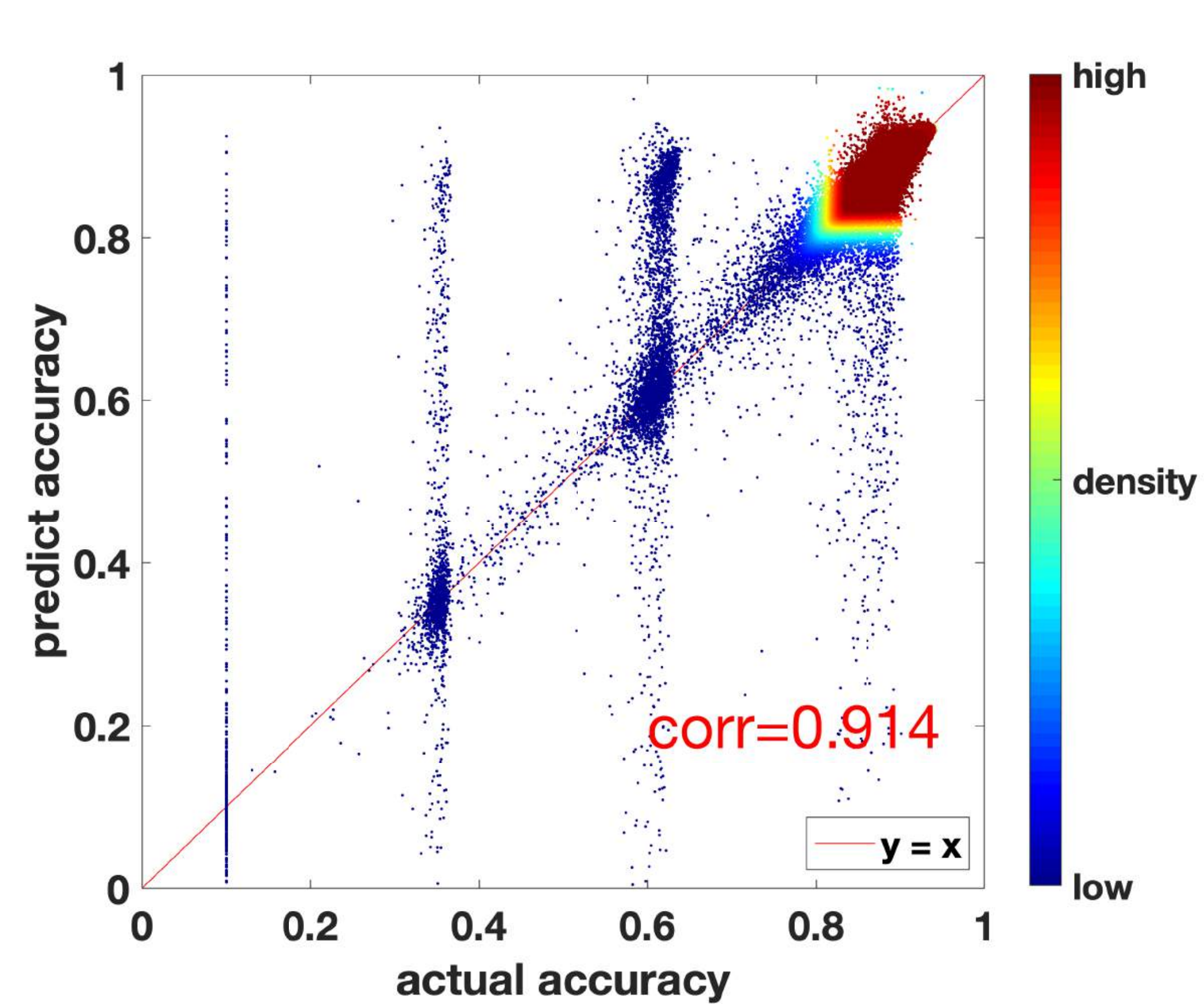}\label{fig:metadnn-rnn-train}}
  \subfloat[][MLP train]{ \includegraphics[width=0.32\columnwidth]{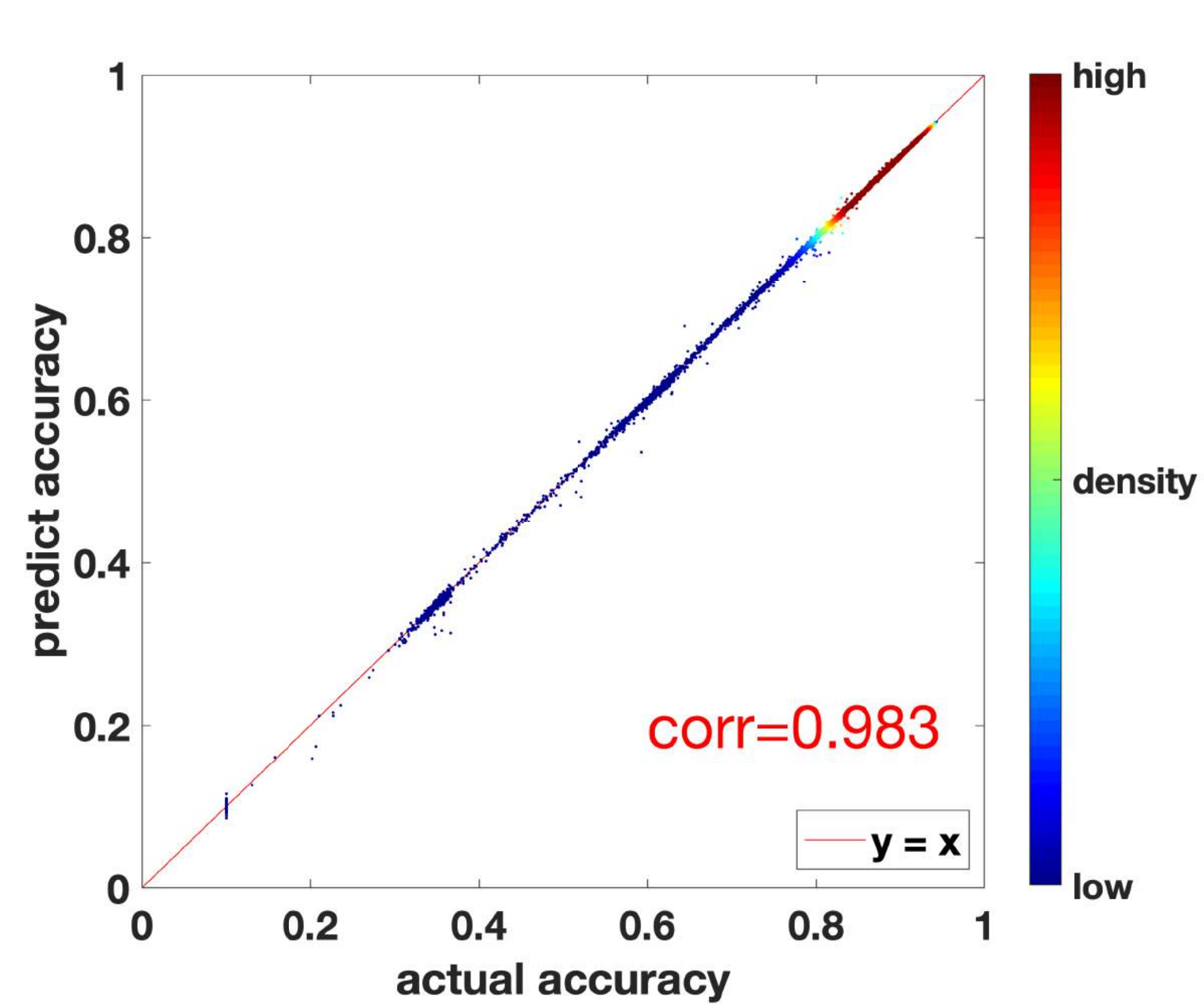}\label{fig:metadnn-mlp-train}} 
  \subfloat[][multi-stage train]{ \includegraphics[width=0.32\columnwidth]{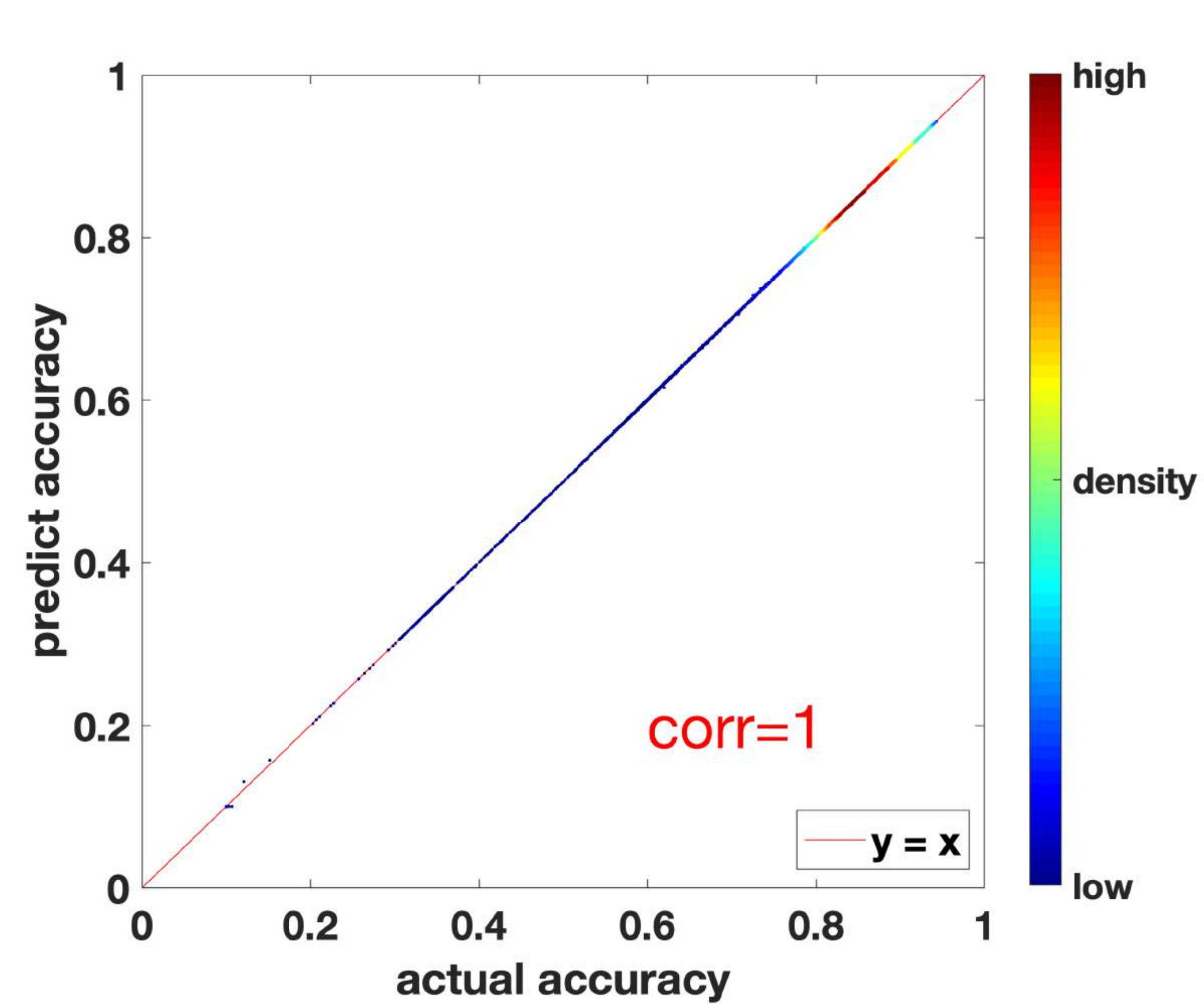}\label{fig:metadnn-mmlp-train}} \\
  \subfloat[][RNN test]{ \includegraphics[width=0.32\columnwidth]{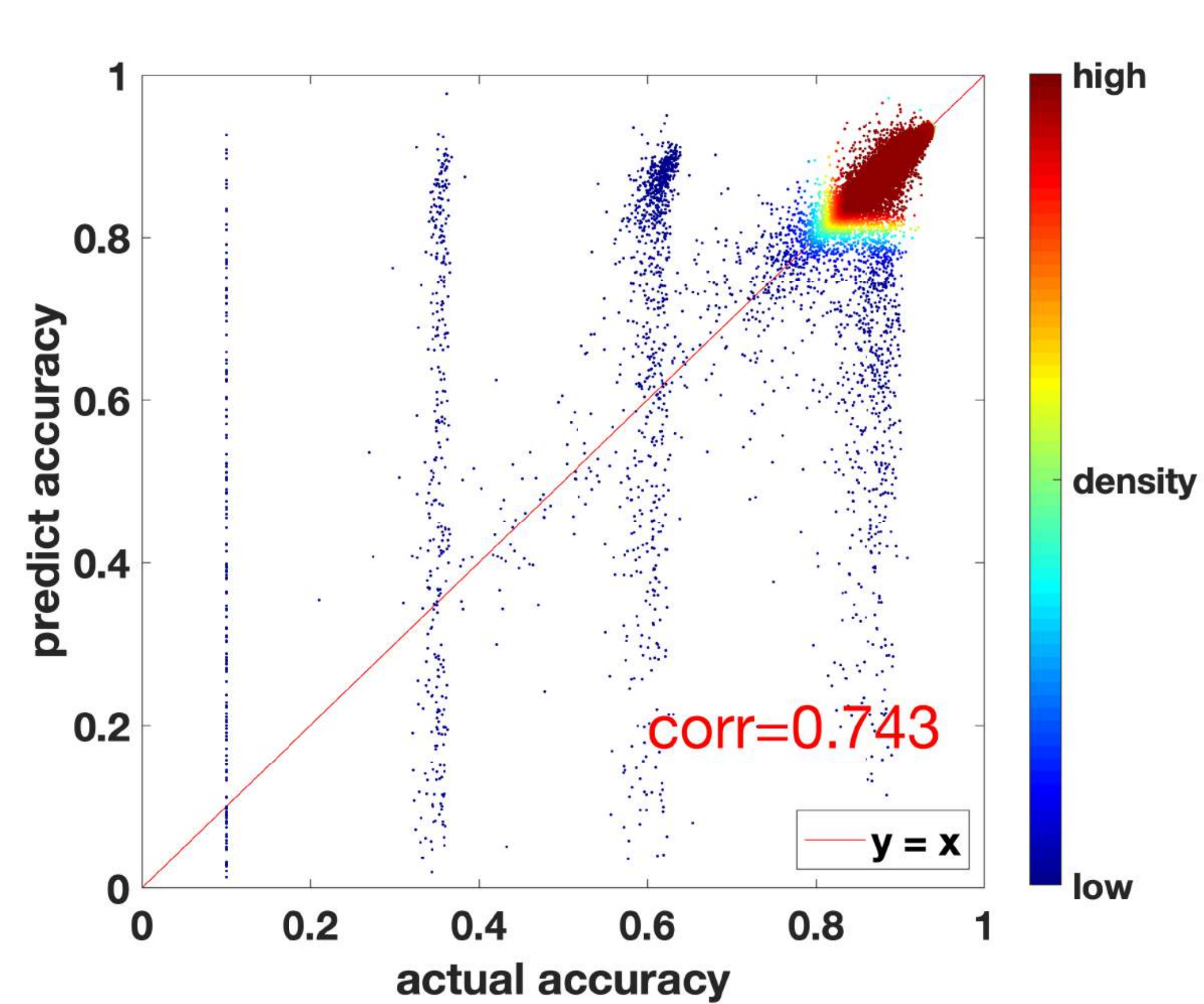}\label{fig:metadnn-rnn-test}} 
  \subfloat[][MLP test]{ \includegraphics[width=0.32\columnwidth]{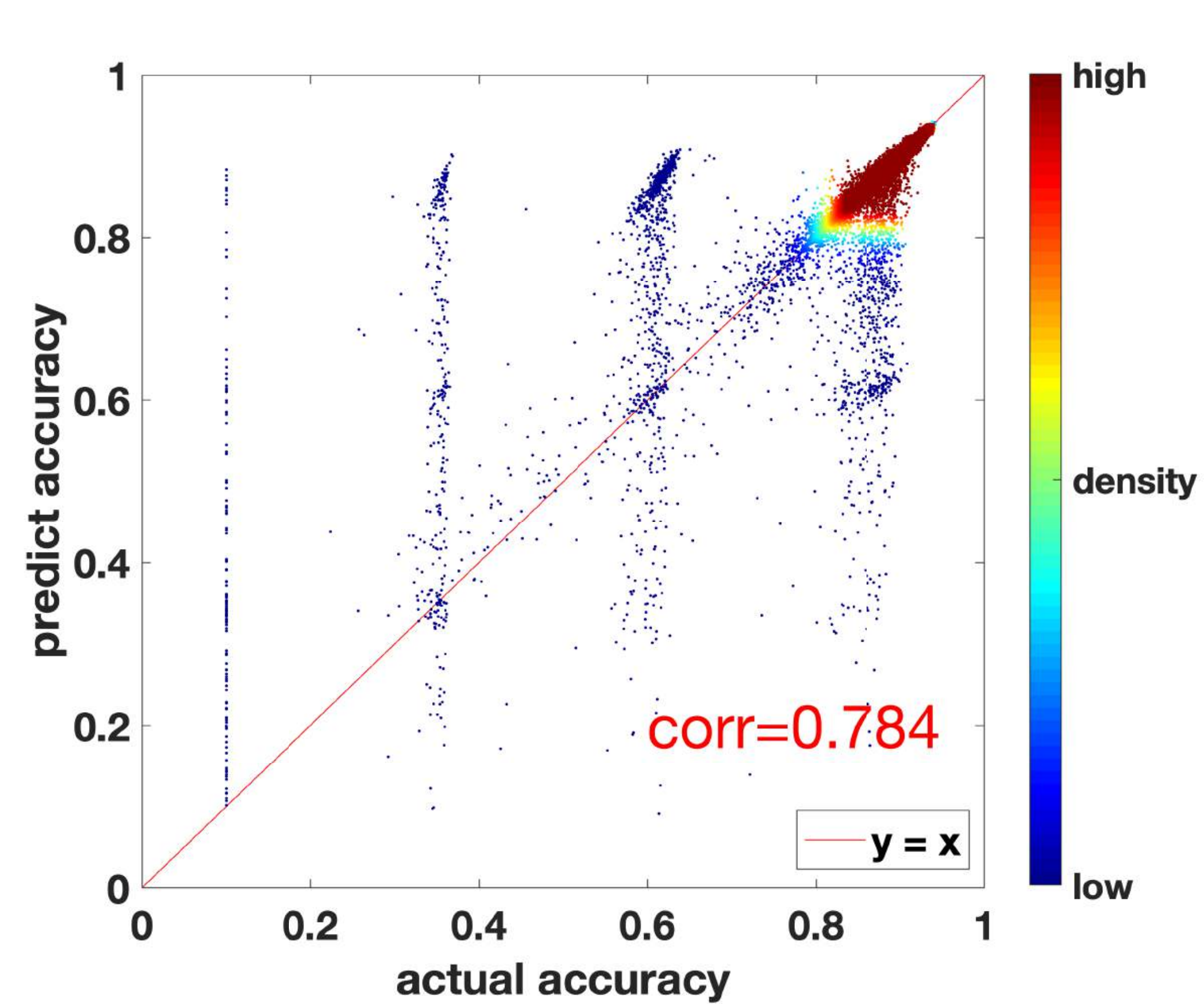}\label{fig:metadnn-mlp-test}}
  \subfloat[][multi-stage test]{ \includegraphics[width=0.32\columnwidth]{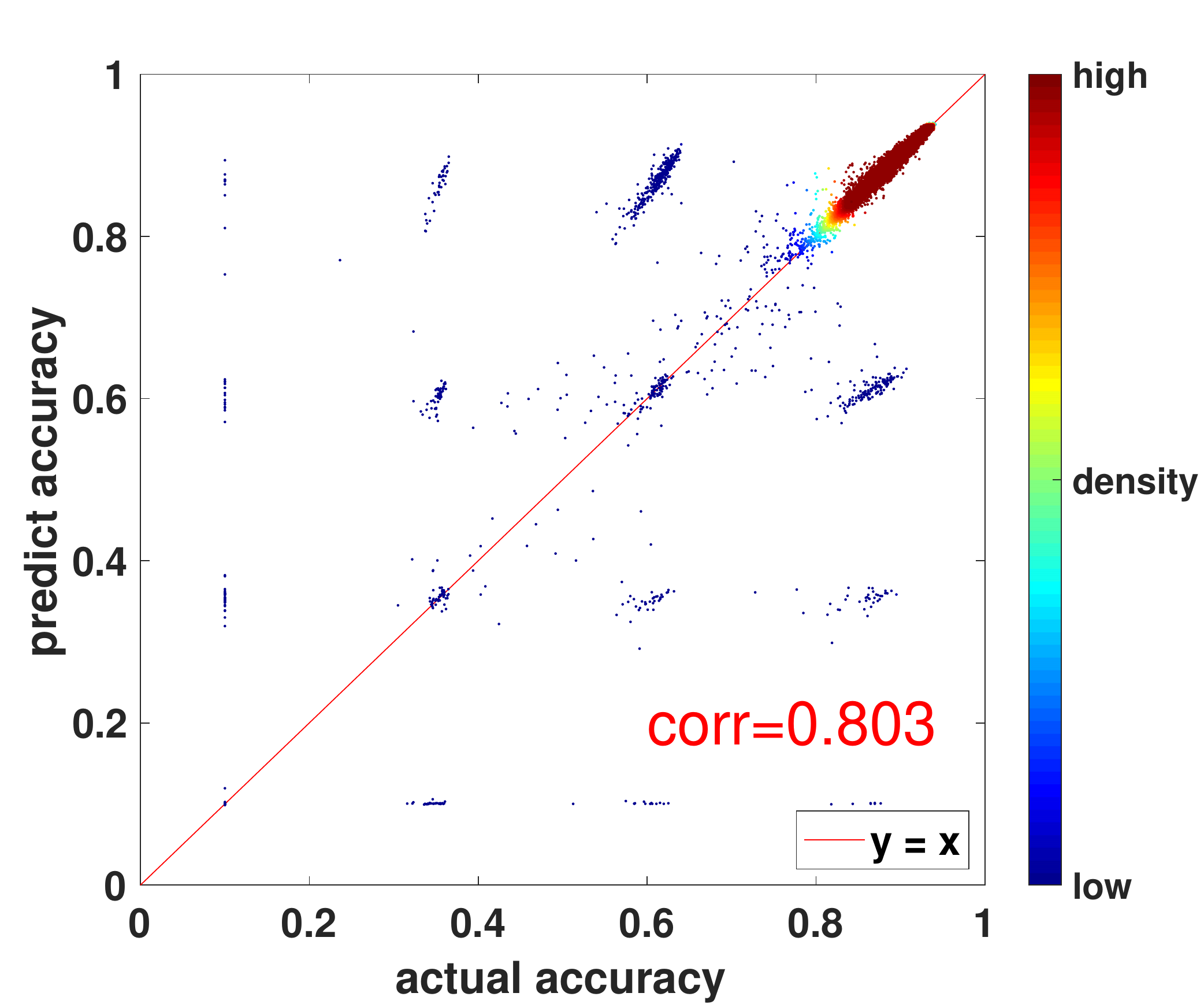}\label{fig:metadnn-mmlp-test}}  \\

  \caption{\textbf{meta-DNN design ablations}: True v.s. predicted accuracies of MLP, RNN and multi-stage MLP on architectures from NASBench. The scatter density is highlighted by color to reflect the data distribution; Red means high density, and blue otherwise. }
  \label{fig:metadnn-design}
  \vspace{-0.2in}
  \end{figure}

\subsection{Meta-DNN Design and its Impact}
\label{sec:metadnn_design}

The metric in evaluating metaDNN is the correlation between the predicted v.s. true accuracy. We used 80\% NASBench for training, and 20\% for testing. Since DNNs have shown great success in modeling complex data, we start with Multilayer Perceptron (MLP) and Recurrent Neural Network (RNN) on building the regression model. Specific architecture details are available in appendix.\ref{ap:metadnn-architecture}.
Fig.~\ref{fig:metadnn-rnn-test} and Fig~.\ref{fig:metadnn-mlp-test} demonstrate the performance of MLP (corr=0.784) is $4\%$ better than RNN (corr=0.743), as the MLP (Fig.~\ref{fig:metadnn-mlp-train}) performs much better than RNN (Fig.~\ref{fig:metadnn-rnn-train}) in the training set. However, MLP still mispredicts many networks around 0.1, 0.4 and 0.6 and 0.8 (x-axis) as shown in Fig.~\ref{fig:metadnn-mlp-test}. This clustering effect is consistent with the architecture distribution in Fig.~\ref{fig:nasbench_acc_dist} for having many networks around these accuracies. To alleviate this issue, we propose a mult-stage model, the core idea of which is to have several dedicated MLPs to predict different ranges of accuracies, e.g. [0, 25\%], along with another MLP to predict which MLP to use in predicting the final accuracy.
Fig.~\ref{fig:metadnn-mmlp-test} shows multi-stage model successfully improves the correlation by 1.2\% from MLP, and the mispredictions have been greatly reduced. Since the multi-stage model has achieved corr = 1 on the training set, we choose it as the backbone regression model for AlphaX. Fig.~\ref{fig:nasbench-results} demonstrates our meta-DNN successfully improves the search efficiency.

\subsection{Transfer Learning}
\begin{figure}
  \begin{center}
    \includegraphics[height=3.3cm]{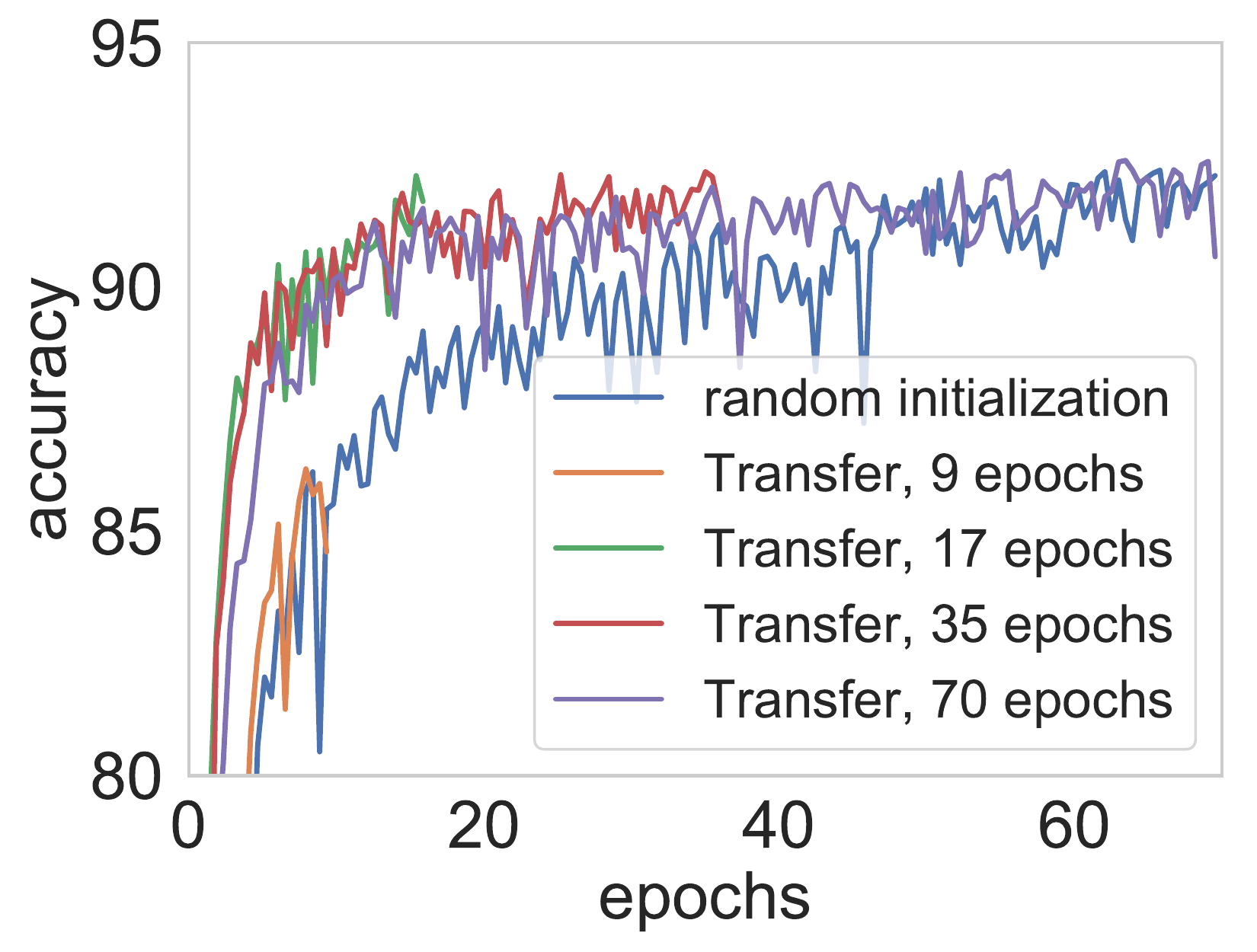}
  \end{center}
  \caption{\textbf{Validation of transfer learning}: transferring weights significantly reduces the number of epochs in reaching the same accuracy of random initializations (Transfer $17\rightarrow70$ epochs v.s. random initialization), but insufficient epochs loses accuracy (Transfer, 9 epochs).}
  \label{fig:transfer_learning}
\end{figure}

The transfer learning significantly speed network evaluations up, and Fig.~\ref{fig:transfer_learning} empirically validates the effectiveness of transfer learning. 
We randomly sampled an architecture as the parent network. On the parent network, we added a block with two new 5x5 separable conv layers on the left and right branch as the child network. We trained the parent network toward 70 epochs and saved its weights. In training the child network, we used weights from the parent network in initializing the child network except for two new conv layers that are randomly initialized. Fig.~\ref{fig:transfer_learning} shows the accuracy progress of transferred child network at different training epochs. The transferred child network retains the same accuracy as training from scratch (random initialization) with much less epochs, but insufficient epochs loses the accuracy. Therefore, we chose 20 epochs in pre-training an architecture if transfer learning applied.  

\subsection{Algorithm Comparisons}
\label{alg-comps-sec}

  \begin{figure}[t]
    \begin{center}
  	  \subfloat[][best acc progression]{ \includegraphics[height=3.3cm]{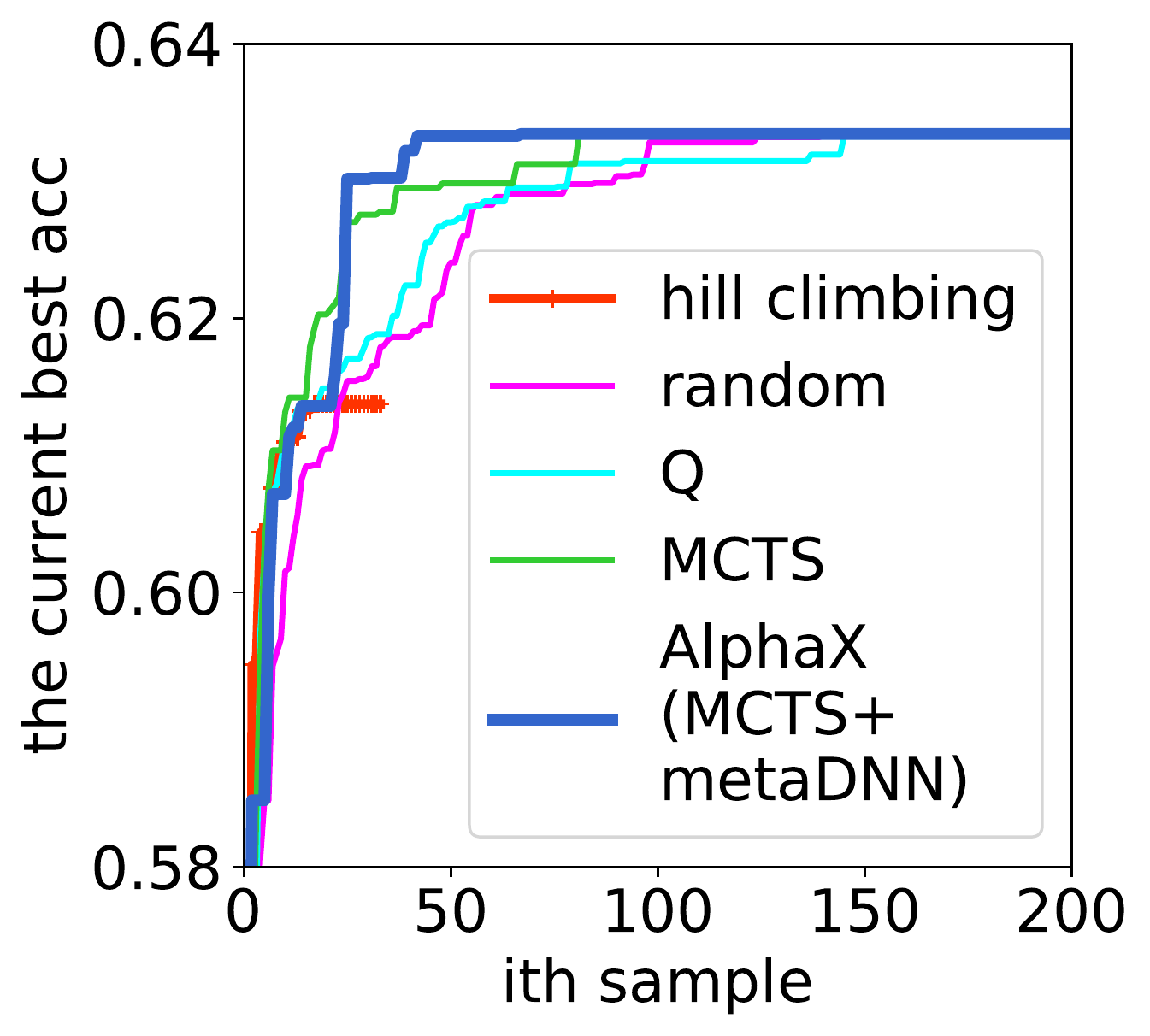}\label{fig:search-comparisions-line}}  \quad
  	\subfloat[][\#samples to the best]{ \includegraphics[height=3.3cm, trim={0.0cm 0cm 0cm 0.0cm}]{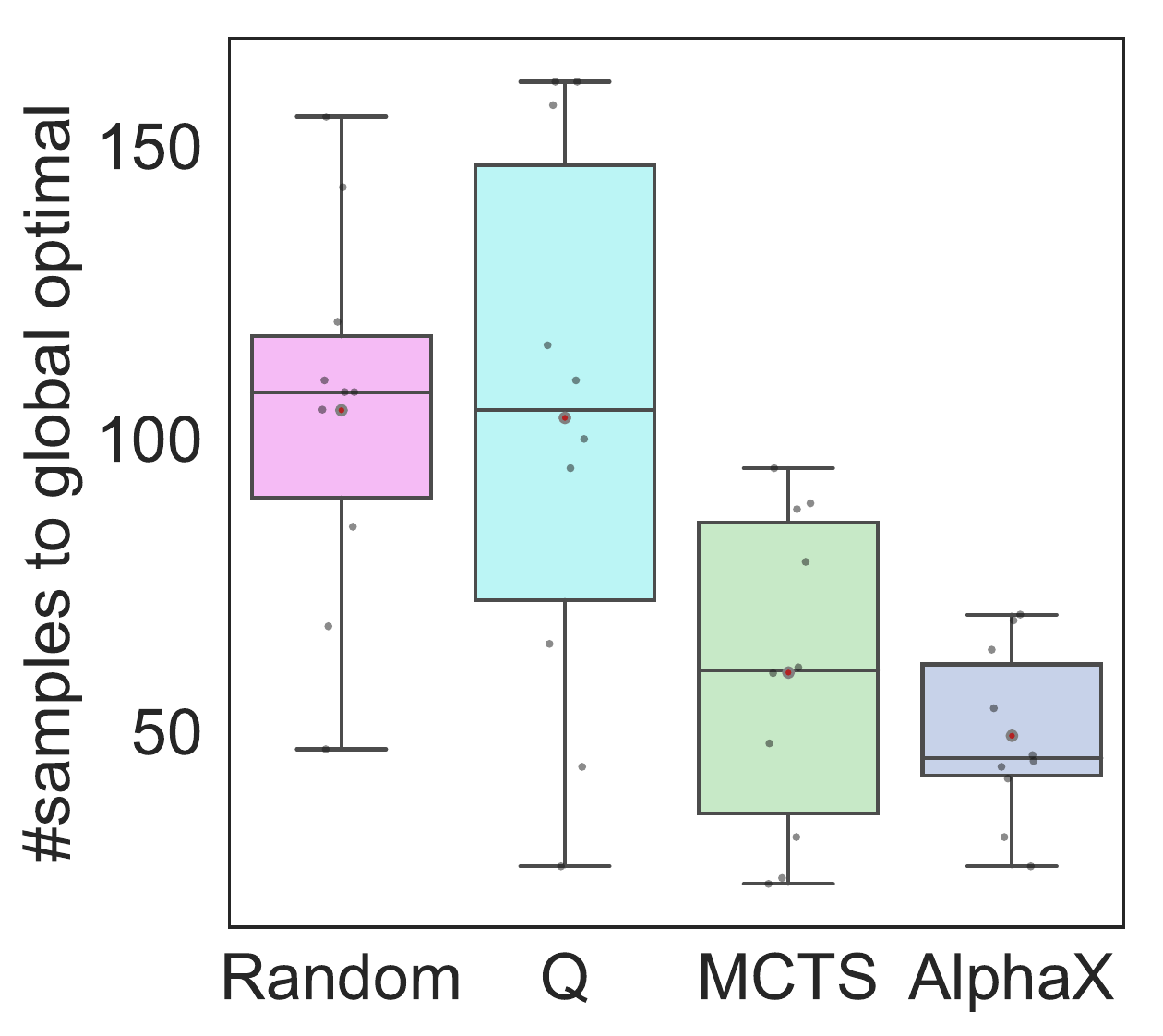}\label{fig:search-comparisions-boxplot}}
    \end{center}
	\vspace{-0.2in}
    \caption{\textbf{Algorithmic comparisions}: AlphaX is consistently the fastest algorithm to reach the global optimal on another simplified search domain (appexdix.\ref{ap:comparison}), while Hill Climbing can easily trap into a local optimal.  }
    \label{fig:alg_comps}
  \end{figure}

Fig.~\ref{fig:alg_comps} evaluates MCTS against Q-Learning (QL), Hill Climbing (HC) and Random Search (RS) on a simplified search space. Setup details and the introduction of the design domain are available in appendix.\ref{ap:comparison}. These algorithms are widely used in NAS \cite{Baker2016, Liu2017, elsken2017simple, greff2017lstm}. We conduct 10 trials for each algorithm. Fig.~\ref{fig:search-comparisions-boxplot} demonstrates AlphaX is 2.3x faster than QL and RS. Though HC is the fastest, Fig.~\ref{fig:search-comparisions-line} indicates HC traps into a local optimal. Interestingly, Fig.~\ref{fig:search-comparisions-boxplot} indicates the inter-quartile range of QL is longer than RS. This is because QL quickly converges to a suboptimal, spending a huge time to escape. This is consistent with Fig.~\ref{fig:search-comparisions-line} that QL converges faster than RS before the 50th samples, but random can easily escape from the local optimal afterward. Fig.~\ref{fig:search-comparisions-boxplot} (MCTS v.s. AlphaX) further corroborates the effectiveness of meta-DNN.

%

\begin{figure}[t]
\begin{center}
\subfloat[][content]{\includegraphics[height=0.6in]{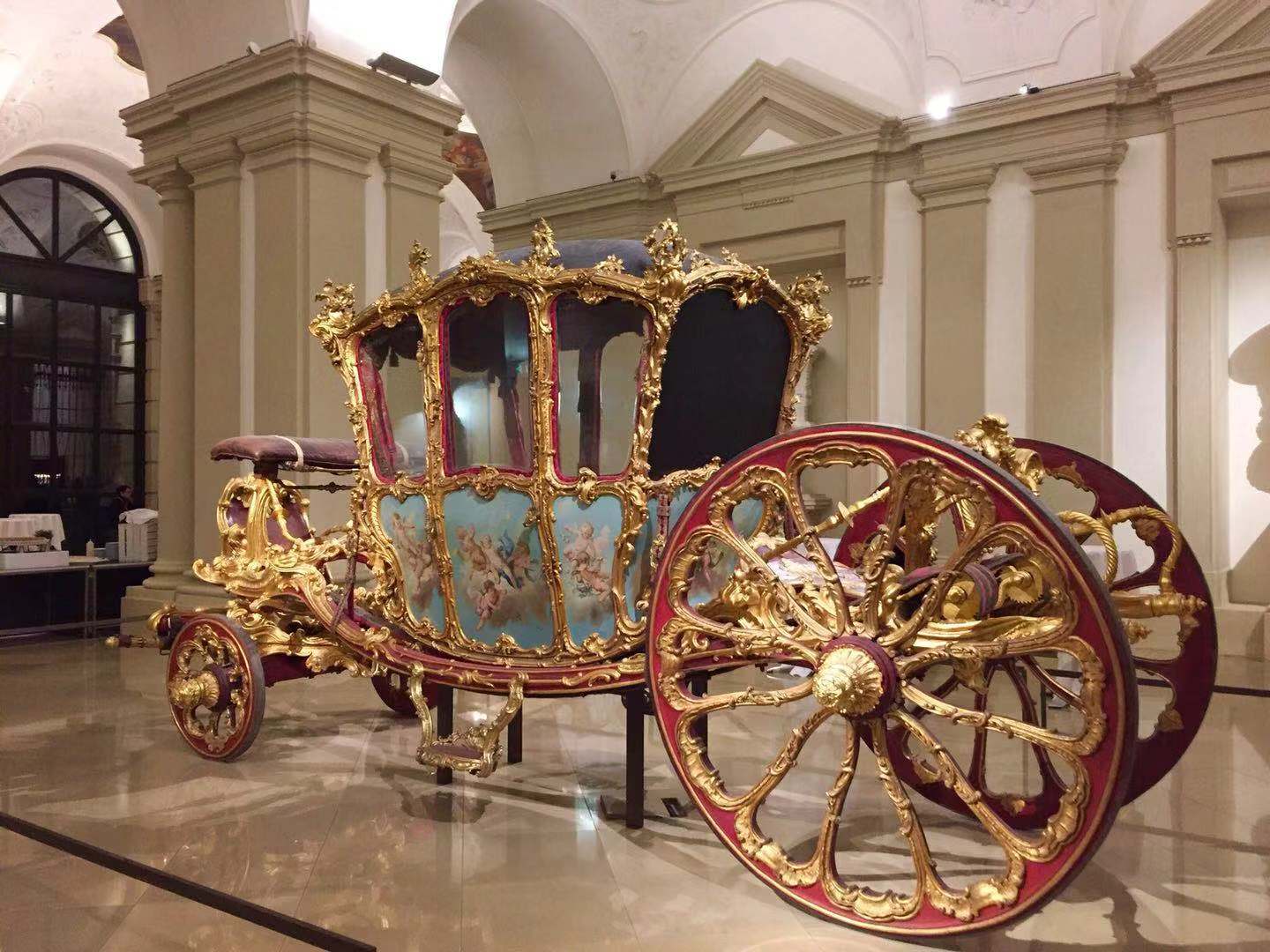}\label{MCTS_state}}
\subfloat[][style]{\includegraphics[height=0.6in]{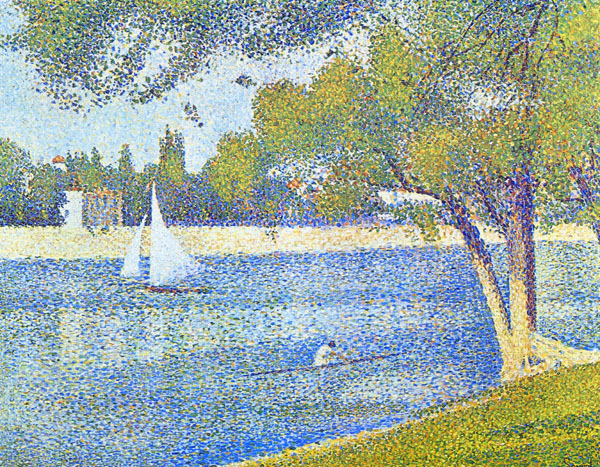}\label{MCTS_state}} 
\subfloat[][VGG]{\includegraphics[height=0.6in]{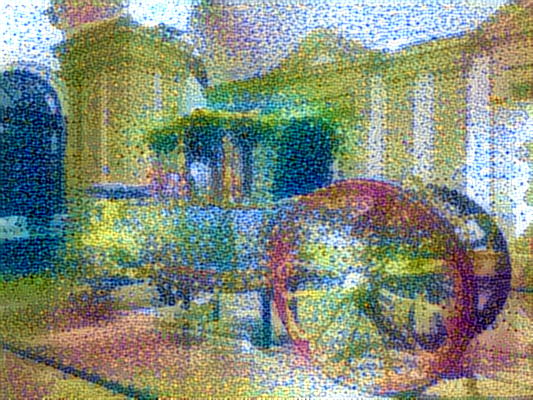}\label{MCTS_no_terminal_state}} 
\subfloat[][AlphaX-1]{\includegraphics[height=0.6in]{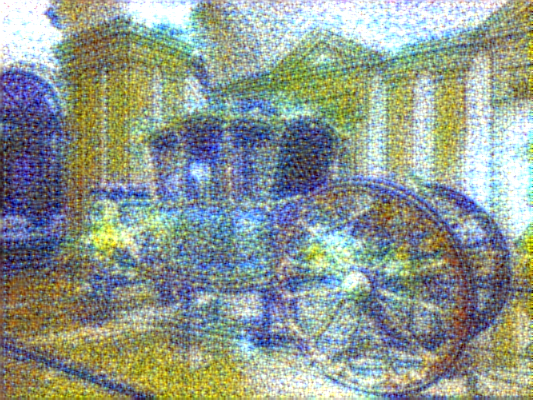}\label{MCTS_no_terminal_state}}
\caption{ \textbf{Neural Style Transfer}: AlphaX-1 v.s. VGG.}
\vspace{-0.2in}
\end{center}
\label{neural-style-transfer}
\vspace{-0.2in}
\end{figure}

\subsection{Improved Features for Applications}

CNN is a common component for Computer Vision (CV) models. Here, we demonstrate the searched architecture can improve a variety of downstream Computer Vision (CV) applications. Please check the Appendix Sec.\ref{ap:setup_vision_models} for the experiment setup. \\
1) \textit{Object Detection}: We replace MobileNet-v1 with AlphaX-1 in SSD \cite{liu2016ssd} object detection model, and the mAP (mini-val) increases from $20.1\%$ to $23.7\%$ at the $300\times300$ resolution. (Fig.\ref{object-detection}) \\
2) \textit{Neural Style Transfer}: AlphaX-1 is better than a shallow network (VGG) in capturing the rich details and textures of a sophisticated style image (Fig.\ref{neural-style-transfer}). \\
3) \textit{Image Captioning}: we replace the VGG with AlphaX-1 in \textit{show attend and tell} \cite{xu2015show}. On the 2014 MSCOCO-val dataset, AlphaX-1 outperforms VGG by 2.4 (RELU-2), 4.4 (RELU-3), 3.7 (RELU-4), respectively (Fig.\ref{fig:captions}).

\section{Conclusion}
We present AlphaX, the first scalable MCTS based design agent for NAS. AlphaX demonstrates superior search efficiency over mainstream algorithms on 3 different search domains, highlighting MCTS as a very promising search algorithm for NAS.

{\small
\bibliographystyle{ieee}
\bibliography{egbib}
}

\newpage

\section{Pseudocode for AlphaX}
\label{ap:pseudocode}

In this section, we describe the pseudocode of the Distributed AlphaX. Algorithm \ref{MCTS} describes the search engine of AlphaX.
Algorithm \ref{server} is the server procedure to send the architecture to train chosen by the MCTS to the client and collect the architectures trained and their scores. Algorithm \ref{client} is the client which trains and tests the architecture provided by the server.

\begin{algorithm}[H]
  \caption{Client}
  \label{client}
    \begin{algorithmic}[1]
      \STATE {\bfseries Require:} Start working once building connection to the server
      \WHILE{True}
      \IF{The client is connected to server}
       \STATE $network \leftarrow \text{Receive}()$
       \STATE $accuracy \leftarrow \text{Train}(network)$
       \STATE Send $(network, accuracy)$ to the Server
      \ELSE
      \STATE Wait for re-connection
      \ENDIF
      \ENDWHILE
    \end{algorithmic}
\end{algorithm}

\begin{algorithm}[H]
  \caption{Server}
  \label{server}
    \begin{algorithmic}[1]
       
      \WHILE{$size(TASK_{-}QUEUE) > 2$}
      \WHILE{no idle client}
      \STATE Continue \Commentl{Wait for dispatching jobs until there are idle clients}
      \ENDWHILE
      \STATE Create a new connection to a random idle client
      \STATE $network \leftarrow TASK_{-}QUEUE.\text{pop}()$
      \STATE Send $network$ to a Client
      \IF{$\text{Received}_{-}\text{Signal}()$}
      \STATE $network, accuracy \leftarrow \text{Receive}_{-}\text{Result}()$
      \STATE $acc(network) \leftarrow accuracy$
      \STATE $state \leftarrow rollout_{-}from(network)$
      \STATE Backpropagation($state$, $(accuracy - \hat{q}(state)) / 2$, 0) \Comment{Replace $\hat{q}$ with $q=Q(s,a)$ in Eq. \ref{reward}}
      \STATE Train the meta-DNN with a new data $(network, accuracy)$
      \ELSE 
      \STATE Continue 
      \ENDIF
      \ENDWHILE
    \end{algorithmic}
\end{algorithm}

\newpage

\begin{algorithm}[H]
  \caption{Search Engine (MCTS)}
  \label{MCTS}
    \begin{algorithmic}[1]
    \STATE \textbf{function} {Expansion}($state$)
    \STATE \quad Create a new node in a tree for state.
    \STATE \quad \textbf{for all} $action$ available at $state$ \textbf{do}
    \STATE \quad $Q(state, action) \leftarrow 0,\quad N(state, action) \leftarrow 0$ 
    \STATE \quad \textbf{end for}
    \STATE \textbf{end function}
    \STATE
    \STATE \textbf{function} {Simulation}($state$)
    \STATE $action \leftarrow none$
    \STATE \quad \textbf{while} $action\;is\;not\;term$ \textbf{do}
    \STATE \quad \quad randomly generate an $action$
    \STATE \quad \quad $next_{-}net$ $\leftarrow$ $\text{Apply}(state, action)$ \Comment{$\text{Apply}$ returns the next state when $action$ is applied to $state$}
    \STATE \quad \textbf{end while}
    \STATE \textbf{end function}
    \STATE
    \STATE \textbf{function} {Backpropagation}($state$, $q$, $n$)
    \STATE \quad \textbf{while} $state\;is\;not\;root$ \textbf{do}
    \STATE \quad \quad $state$ $\leftarrow$ $\text{parent}(state)$
    \STATE \quad \quad $Q(state, action) \leftarrow Q(state, action) + q$
    \STATE \quad \quad $N(state, action)$ $\leftarrow$ $N(state, action) + n$
    \STATE \quad \textbf{end while} 
    \STATE \textbf{end function}
    \STATE
      \STATE {\bfseries Require:} Start from the root
      \WHILE {$episode < MAX_{-}episode$}
      \STATE $\text{Server}()$  
      \STATE $cur_{-}state$ $\leftarrow$ $root_{-}node$
      \STATE $i \leftarrow 0$
      \WHILE{$i < MAX_{-}tree_{-}depth$}
      \STATE $i \leftarrow i + 1$
      \STATE $next_{-}action$ $\leftarrow$ $\text{Selection}(cur_{-}state)$ \Commentl{Select an action based on Eq. \ref{ucb1_update}}
      \IF{$next_{-}state$ not in tree}
      \STATE $next_{-}state \leftarrow \text{Expansion}(next_{-}action)$
      \STATE $T_t \leftarrow \text{Simulation}_t(next_{-}state)$ for $t=0...k$ \Comment{$k$ is the number of simulations we run using the Meta-DNN}
      \STATE $TASK_{-}QUEUE.\text{push}(T_0)$
      \STATE $rollout_{-}from(T_0) \leftarrow next_{-}state$
      \STATE $\hat{q}(next_{-}state) \leftarrow \frac{1}{k} \sum_{i=1..k}Pred(T_i)$ \Comment{$Pred$ returns an accuracy predicted by the Meta-DNN}
      \STATE Backpropagation$(next_{-}state, \hat{q})$ \Commentl{Preemptive backpropagation to send $\hat{q}$}
      \ENDIF
      \ENDWHILE
      \ENDWHILE
    \end{algorithmic}
\end{algorithm}


\newpage

\begin{table*}[t]
  \small
  \caption{The code of different types of layers}
  \label{app:layers_used}
  \centering
  \begin{tabular}{ l l l l l l l l}
    \toprule
          layers & code & layer & code & layer & code & layer & code \\
    \midrule
    	  3x3 avg pool & 1 & 3x3 max pool & 4 & 3x3 conv & 7 & 3x3 depth-separable conv & 10  \\
    	  5x5 avg pool & 2 & 5x5 max pool & 5 & 5x5 conv & 8 & 5x5 depth-separable conv & 11  \\
    	  7x7 avg pool & 3 & 7x7 max pool & 6 & identity & 9 & 7x7 depth-separable conv & 12  \\
    \bottomrule
    \label{layers_used}
  \end{tabular}
\vspace{-0.3in}
\end{table*}

\label{ap:arch}

\begin{figure}
  \begin{center}
    \includegraphics[width=0.49\columnwidth, height=1.6in]{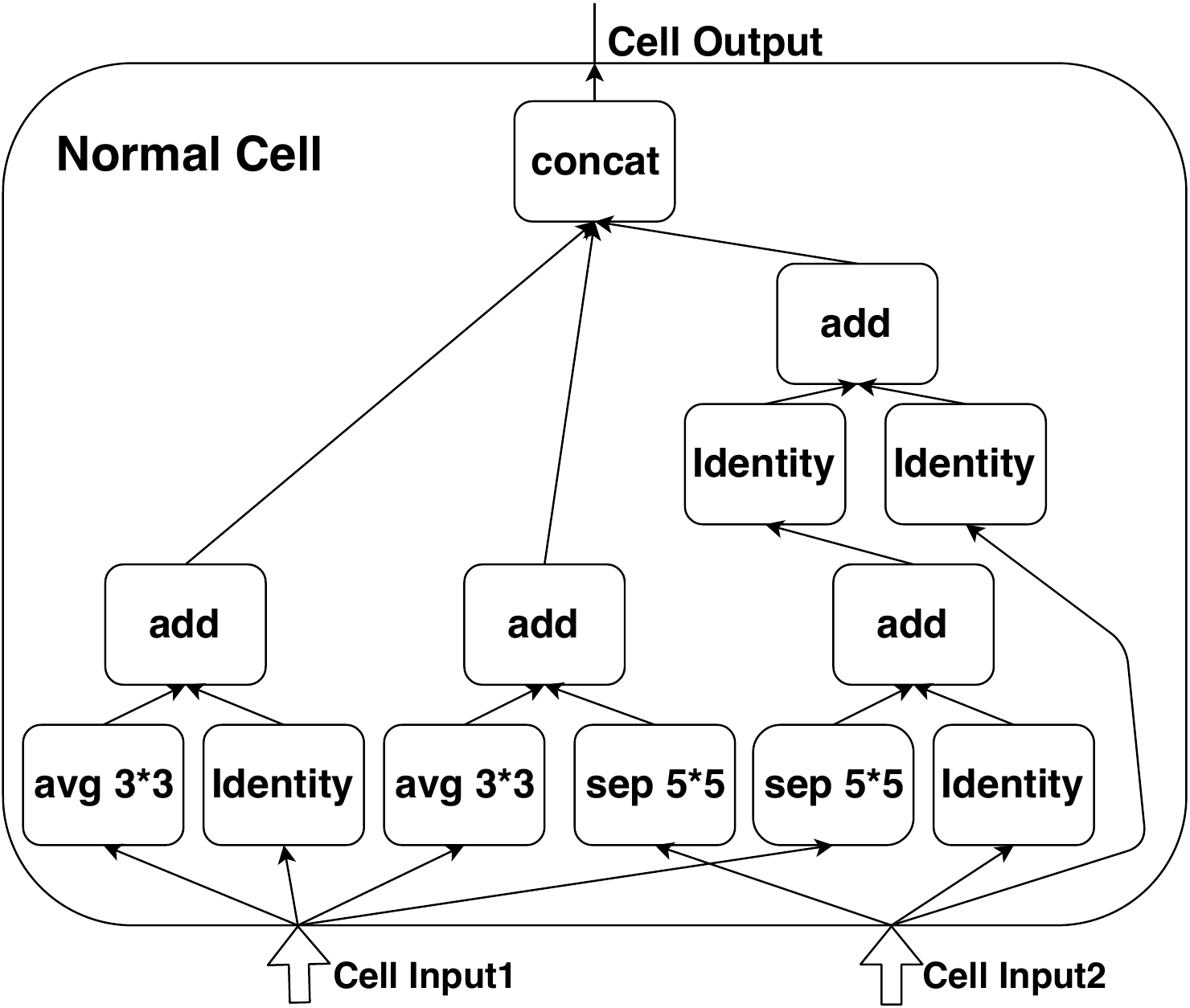}
    \includegraphics[width=0.49\columnwidth, height=1.6in]{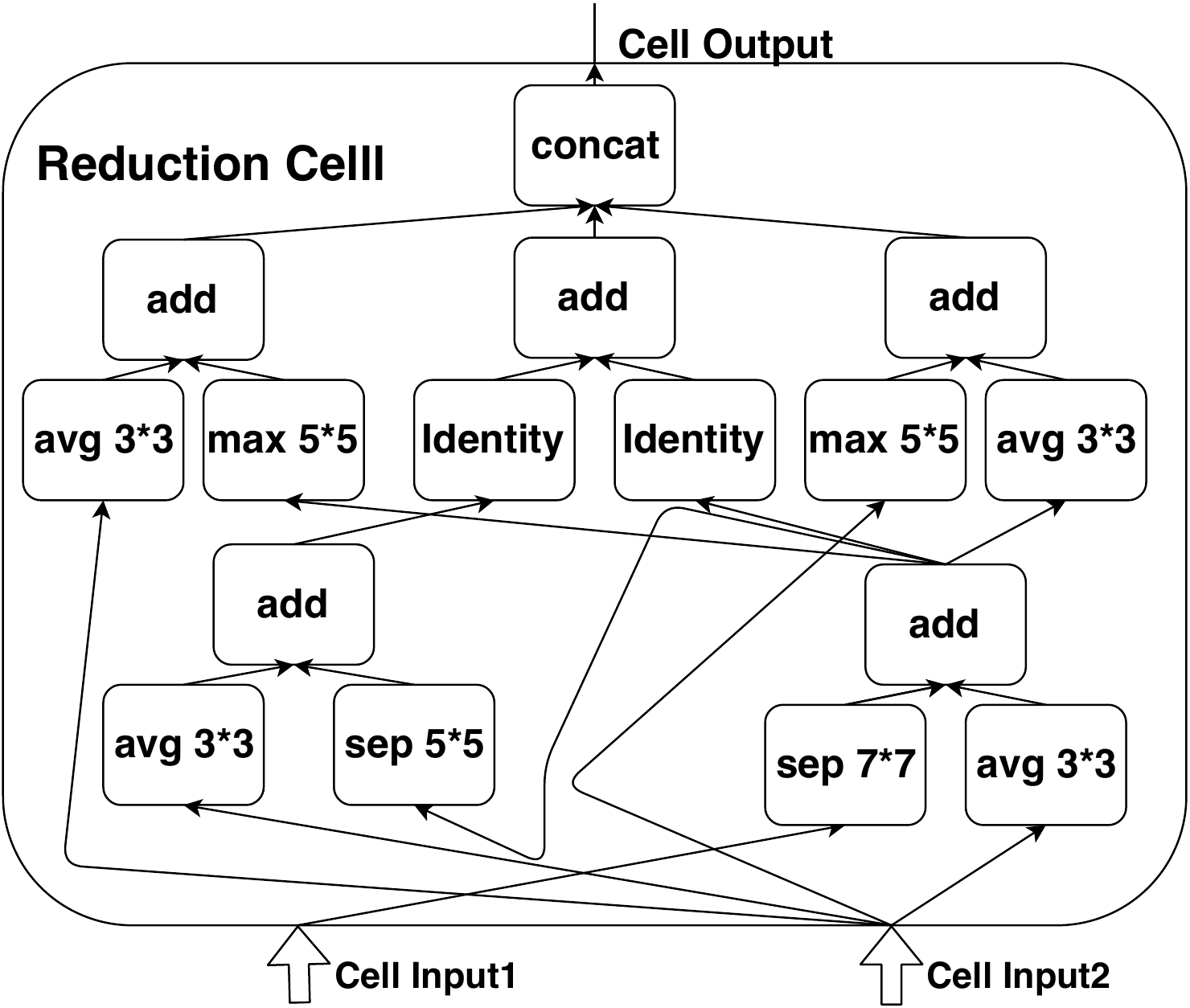}
  \end{center}
  \caption{the $RCell$ and $NCell$ of AlphaX-2}
  \label{ap:3382}
\end{figure}

\section{Details of the State and Action Space}
\label{ap:state_space}

This section contains the description of the state and action space for NASNet design space.

We constrain the state space to make the design problem manageable. The state space exponentially grows with the depth: a k layers linear network has $n^{k}$ architecture variations, where n is the number of layer types. We leverage the GPU DRAM size, our current computing resources and the design heuristics from leading DNNs, to propose the following constraints on the state space: 1) a branch has at most 1 layer; 2) a cell has at most 5 blocks; 3) the depth of blocks is limited to 2; 5) we use the layers in TABLE.\ref{layers_used}:


%
%
%
%

Actions also preserve the constraints imposed on the state space. If the next state reaches out of the design boundary, the agent automatically removes the action
from the action set. For example, we exclude the "adding a new layer" action for a branch if it already has 1 layer. So, the action set is dynamically changing w.r.t states.




\section{Experimental Setup for Section \ref{alg-comps-sec}}
\label{ap:comparison}

The state space we consider consists of small simple CNNs.
We consider a convolution layer and a softmax layer.
For a convolutional layer, we allow a range of 1 or 2 for stride, 32 or 64 for filters, and 2 or 4 for kernels.
We set the maximum depth to 3.
We constraint that the final layer is always a dense layer with output size set to the number of classes.

Actions consist of (1) add conv layer, (2) add softmax layer, (3) increment or decrement one of the parameters for a convolution layer in the current CNN.
For MCTS, random, and Q-learning agents have a terminal action to terminate the episode.

\textit{MCTS with meta-DNN}: We implemented MCTS search algorithm followed the procedure with \ref{sec:search}. $c$ from Eq.\ref{ucb1_update} is set to 200. The design of meta-DNN is consistent with \ref{sec:metadnn}. The meta-DNN model uses SGD optimizer with 0.9 momentum rate. All parameters in fully connected layers are initialized with the random distribution. The learning rate is set to 0.0001.

\textit{MCTS without meta-DNN}: we also present the results without meta-DNN, the experiment setup is consistent with above but without meta-DNN assisted simulation.

\textit{Random}: agent selects action uniformly at random.

\textit{Q-learning}: We implemented a tabular Q-learning agent with $\epsilon$-greedy strategy.
The learning rate is set to 0.2. We set the discount factor to be 1 in order not to prioritize short-term rewards. We fix $\epsilon$ to 0.2.
We initialize the Q-value with 0.5.

\textit{Hill Climbing}: For a hill climbing, an agent starts from a randomly chosen initial state. It trains every architecture in the child nodes and moves to the child node of which architecture performed the best, and repeat this procedure. Unlike MCTS and Q-learning which trains a NN only when it is a terminal state, hill climbing considers every state (and its child nodes) it visits to train. As such, we do not have a terminal action for hill climbing.
As we observed that the hill climbing tends to stick to a local optimum, we restart from a randomly chosen initial state if it visits the same state twice in the trajectory.

\section{Experiment Setup for Searching and Training}
\label{ap:experiment_setup}

\subsection{Setup for searching networks on NASBench}

\textit{Regularized Evolution}: We implemented Regularized Evolution\cite{real2018regularized} search algorithm for searching the best architecture on NASBench. The population size is set to 500. The tournament size is set to 50. We only mutate the best architecture in tournament set and replace the oldest individual in the population. Once we find the best architecture on NASBench set, we terminate the search process.

\textit{AlphaX}: We implemented MCTS search algorithm followed the procedure with \ref{sec:search} on NASBench. $c$ from Eq.\ref{ucb1_update} is set to 2. The simulation times $k$ from Eq.\ref{reward}  is set to 10. The design of meta-DNN is consistent with \ref{sec:metadnn_design}. The setup for meta-DNN is consistent with \ref{ap:metadnn-architecture}.

\subsection{Setup for searching networks on CIFAR}
\label{ap:mcts_setup_cifar}
We use 16 NV-1080ti gpus for searching procedure. One of them is the server running the searching program and remaining 15 gpus are clients for training the searched architectures. We use dictionary data structure in Python to save the searched architectures and convert to json file to store them in the disk. All of our training procedures are implemented in MXNET framework.
The setup for training CIFAR-10 during the search are as follows: 1) We early terminate the training at the 70th epoch(3 periods of cosine restart learning rate schedule\cite{cosine_restart}) due to the limited computing resources; then we rank networks to filter out top ones to perform additional 560 epochs to acquire the final accuracy.
2) Cutout is applied \cite{zoph2017learning} by using 1 crop of size $16\times16$. 3) Our models use cosine restart learning rate schedule\cite{cosine_restart} with 3 periods, the base learning rate is 0.05 and the batch size is 144. 4) We use the momentum optimizer with momentum rate set to 0.9 and L2 weight decay. 5) We also use dropout ratio schedule in the training. The droppath ratio is set to 0.3 and dense dropout ratio is set to 0.2 for searching procedure and applied \textit{ScheduleDropPath}\cite{zoph2017learning} for the final training. 6) We use an auxiliary classifier located at 2/3 of depth of the network. The loss of the auxiliary classifier is weighted by 0.4\cite{SzegedyVISW15}. 7) The weights of our models are initialized with Gaussian distribution subjected to 0.01 standard deviation. 8) we randomly crop 32x32
patches from upsampled images of size $40\times40$ and apply random horizontal flips consistent with\cite{zoph2017learning}. 

\subsection{Setup for ImageNet}
\label{ap:setup_img}
The setup for training ImageNet are as follows: 1) We construct the network for ImageNet with searched $RCell$ and $NCell$ according to Fig.\ref{best-arch} in \cite{zoph2017learning}. 2) The input image size is $224\times224$ (the mobile setting across literature). 3) Our models for ImageNet use polynomial learning rate schedule, starting with 0.05 and decay through 200 epochs. 4) We use the momentum optimizer with momentum rate set to 0.9 and L2 weight decay. 5) Our model uses an auxiliary classifier located at 2/3 of depth of the network. The loss of the auxiliary classifier is weighted by 0.4\cite{SzegedyVISW15}. 6) Dense dropout is applied to the final softmax layer with probability 0.5. 7) We set the batch size as 256. 8) The weights of our models are initialized with Gaussian distribution subjected to 0.01 standard deviation.

\begin{figure}[t]
  \begin{center}
    \subfloat[][RNN predictor]{\includegraphics[height=1.4in]{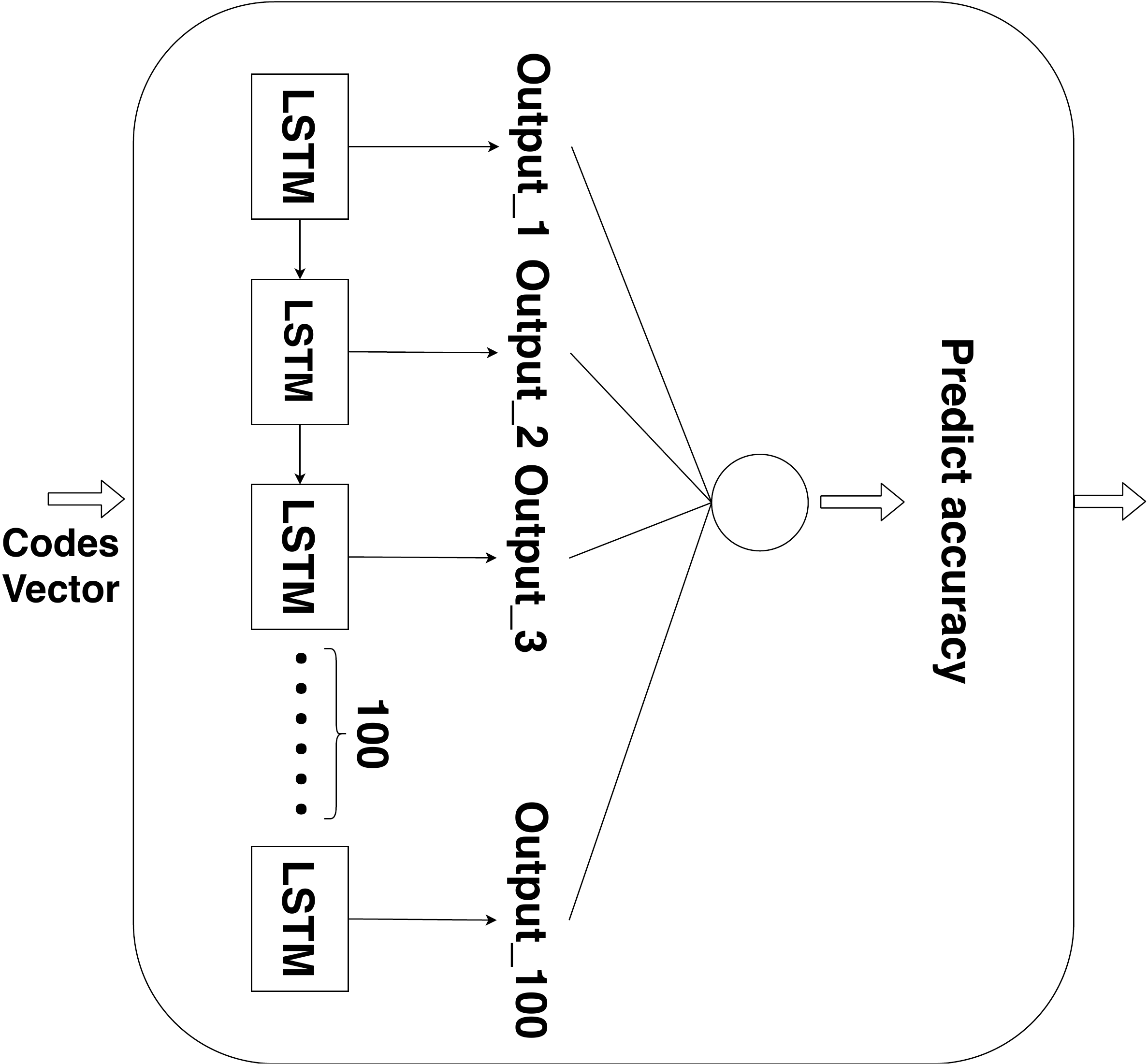}\label{meta_dnn_rnn}} \quad
    \subfloat[][MLP predictor]{\includegraphics[height=1.4in]{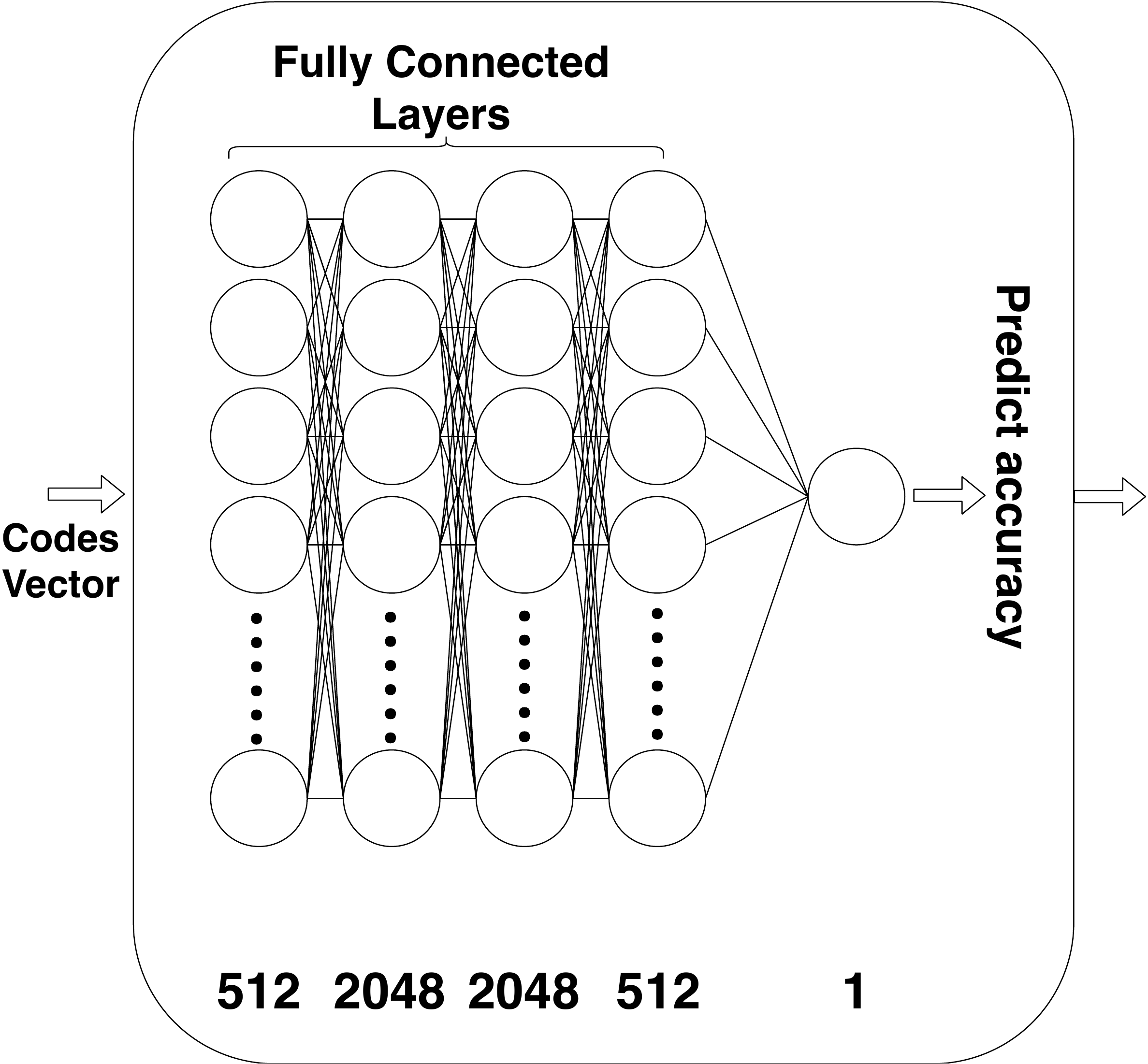}\label{meta_dnn_mlp}} 
  \end{center}
\caption{ \textbf{Architecture of metaDNN}: (a) the architecture of recurrent neural network style metaDNN (b) the architecture of multilayer perceptron style metaDNN. }
\label{meda_dnn}
\end{figure}

\section{MetaDNN Architecture and Setup}

\label{ap:metadnn-architecture}

\subsection{Architecture and setup for recurrent neural network (RNN)}
We use an LSTM model as our predictor which is consistent with \cite{liu2017progressive}. The hidden state size is set to 100. The embedding size is set to 100 as well. The final LSTM hidden state goes through a fully-connected layer and sigmoid to regress the validation accuracy. Fig.\ref{meta_dnn_rnn} shows the architecture of the RNN predictor. 

The setup for training RNN predictor are as follows: 1) There are total of 20 epochs for each training. 2) The base learning late is set to 0.00002 and the batch size is 128. 3) We use the Adam optimizer for training. 4) The embeddings use uniform initialization in range [-0.1, 0.1]. the weights of fully connected layers are initialized with Uniform distribution with initial bias 0.

\subsection{Architecture and setup for multilayer perceptron (MLP)}

We implement both multi-stage MLP model and single MLP to predict the network accuracy. Both of multi-stage model and single MLP model share the same architecture. We use 5 fully connected layers with 512, 2048, 2048, 512 and 1 nodes. The final fully-connected layer uses the sigmoid function to regress the validation accuracy. Fig.\ref{meta_dnn_mlp} shows the architecture of the MLP predictor. 

The setup for training MLP predictor are as follows: 1) There are total of 20 epochs for each training. 2) The base learning late is set to 0.00002 and the batch size is 128. 3) We use the Adam optimizer for training.
4) The weights of our models are initialized with Uniform distribution with initial bias 0.

\begin{figure}[t]
\centering 
\includegraphics[height=2.5in, width=0.15\textwidth]{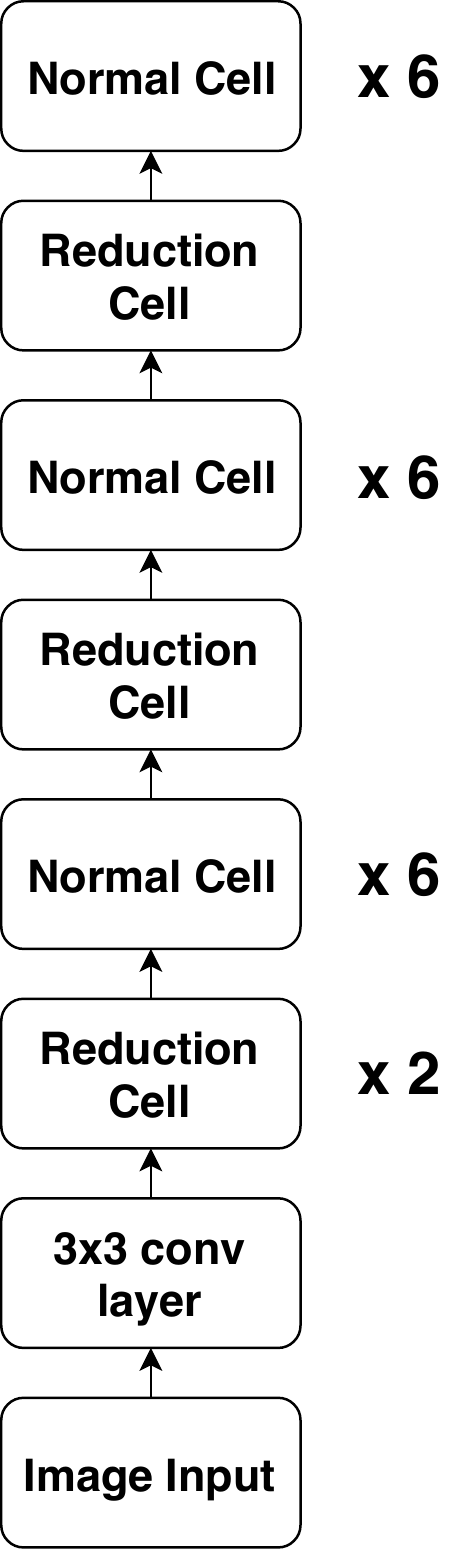}
\caption{Cell based architecture of AlphaX model}
\label{neural_style}
\end{figure}

\section{Setup for Vision Models}

\begin{figure*}[!htb]
\vspace{-0.4in}
\minipage{\textwidth}
	\centering
	\subfloat[][AlphaX-1]{\includegraphics[height=1.4in, trim={6cm 8cm 0cm 5cm},clip]{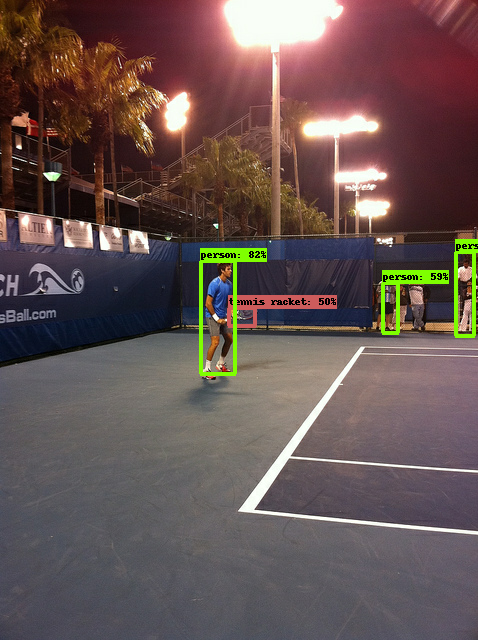}\label{MCTS_state}} \quad
	\subfloat[][MobileNet-v1]{\includegraphics[height=1.4in, trim={6cm 8cm 0cm 5cm},clip]{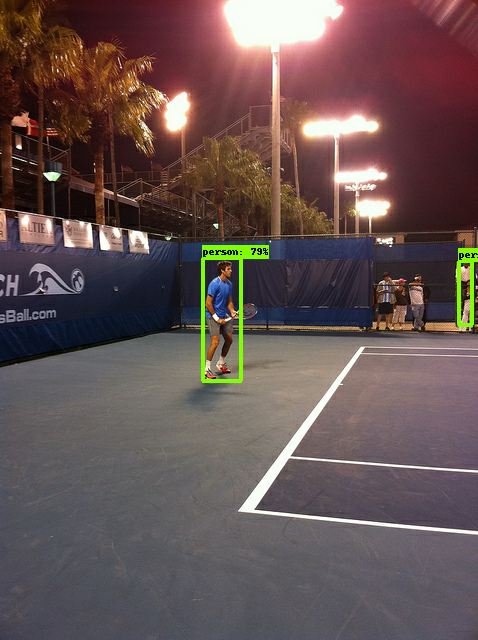}\label{MCTS_state}} \quad
	\subfloat[][AlphaX-1]{\includegraphics[height=1.4in, trim={6cm 3cm 6cm 4cm},clip]{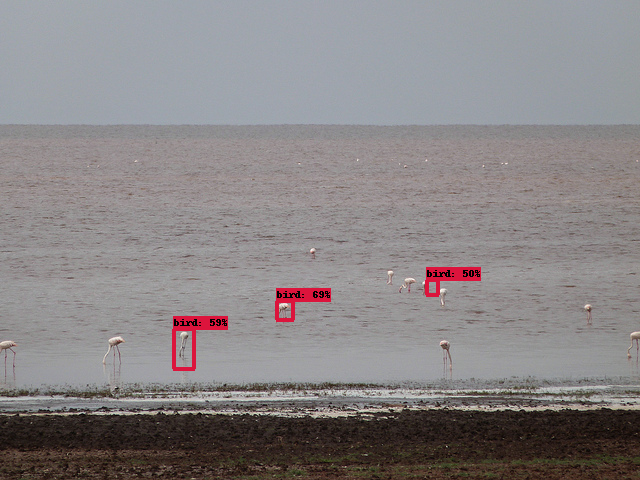}\label{MCTS_state}} \quad
	\subfloat[][MobileNet-v1]{\includegraphics[height=1.4in, trim={6cm 3cm 6cm 4cm},clip]{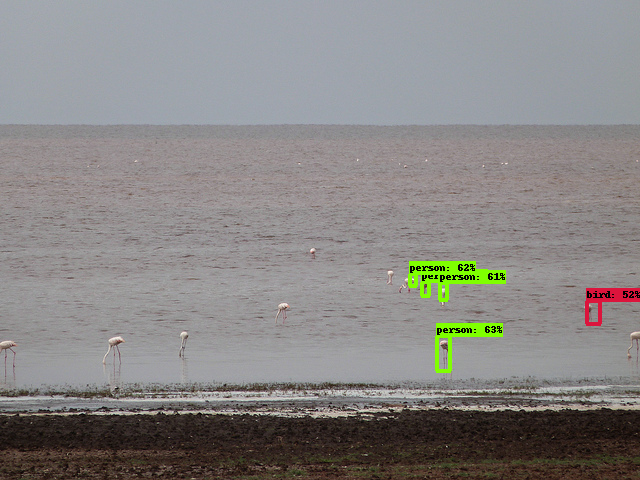}\label{MCTS_state}} \quad  \\
	\caption{\textit{Object Detection}: the object detection system is more precise with AlphaX-1 than MobileNet. }
	\label{object-detection}
\endminipage\hfill
\minipage{\textwidth}
	\centering
	\subfloat[][content]{\includegraphics[height=1.2in]{fig/neural-style-content.jpeg}\label{MCTS_state}} \quad
	\subfloat[][style]{\includegraphics[height=1.2in]{fig/neural-style-style.jpeg}\label{MCTS_state}} \quad 
	\subfloat[][VGG synthesized picture]{\includegraphics[height=1.2in]{fig/neural-style-vgg.png}\label{MCTS_no_terminal_state}}  \quad
	\subfloat[][AlphaX-1 synthesized picture]{\includegraphics[height=1.2in]{fig/neural-style-alphax.png}\label{MCTS_no_terminal_state}}  \quad \\
	\caption{ \textit{Neural Style Transfer}: AlphaX-1 is better than VGG in capturing the style with sophisticated details and textures.}
	\label{neural-style-transfer}
\endminipage\hfill
\minipage{\textwidth}%
	\centering
	\subfloat[][VGG: a tennis player is playing tennis on the court. \\
				AlphaX-1: a couple of people playing tennis on a tennis court. 				]{\includegraphics[height=1in]{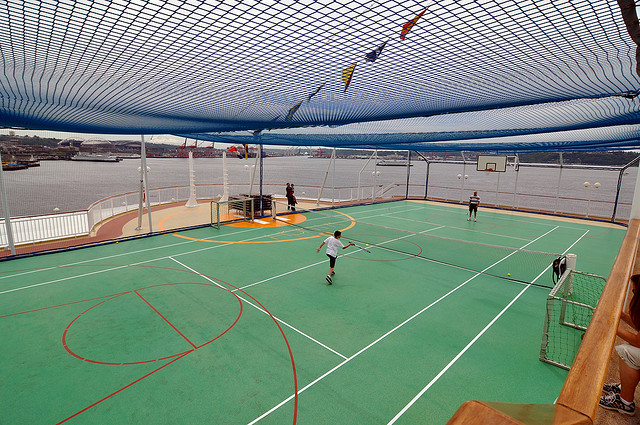}\label{MCTS_state}} \quad
	\subfloat[][VGG: a bus is parked on the side of the road. \\ AlphaX-1: a blue bus is driving down a street.]{\includegraphics[height=1in]{fig/image-caption/COCO_val2014_000000133819.jpg}\label{MCTS_state}} \quad
	\subfloat[][VGG: a cup of coffee and a cup of coffee. \\
	AlphaX-1: a book and a cup of coffee on a table. ]{\includegraphics[height=1in]{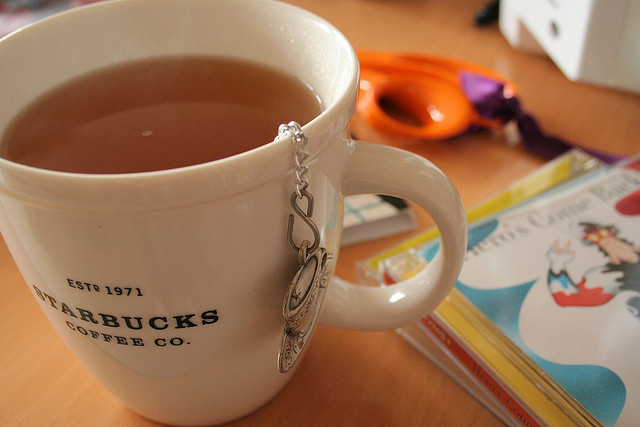}\label{MCTS_state}} \quad
	\subfloat[][VGG: a fire hydrant on the side of a road.\\
	AlphaX-1: a red fire hydrant on the side of a road.
	]{\includegraphics[height=1in]{fig/image-caption/COCO_val2014_000000398097.jpg}\label{MCTS_state}}\quad
	\caption{ \textit{Image Captioning}: AlphaX-1 captures more details than VGG in the captions.}
	\label{fig:captions}
\endminipage
\end{figure*}

\label{ap:setup_vision_models}
\subsection{Object detection}
We use AlphaX-1 model pre-trained on ImageNet dataset. The training dataset is MSCOCO for object detection\cite{LinMBHPRDZ14} which contains 90 classes of objects. Each image is scaled to $300\times300$ in RGB channels. We trained the model with 200k iterations with 0.04 initial learning rate and the batch size is set to 24. We applied the exponential learning rate decay schedule with the 0.95 decay factor. Our model uses momentum optimizer with momentum rate set to 0.9. We also use the L2 weight decay for training. We process each image with random horizontal flip and random crop\cite{liu2016ssd}. We set the matched threshold to 0.5, which means only the probability of an object over 0.5 is effective to appear on the image. We use 8000 subsets of validation images in MSCOCO validation set and report the mean average precision (mAP) as computed with the standard COCO metric library\cite{MSCOCO2014}. 

\subsection{Neural style}
We implement the neural style transfer application by replacing the VGG model to AlplaX-1 model\cite{neural_style}. AlplaX-1 model is pre-trained on ImageNet dataset. In order to produce a nice result, we set the total 1000 iterations with 0.1 learning rate. We set 10 as the style weight which represents the extent of style reconstruction and 0.025 as the content weight which represents the extent of content reconstruction. We test different kinds of combinations of the outputs of different layers. Fig.\ref{neural_style} shows the structure of AlphaX model, we found that for AlphaX-1 model, the best result can be generated by the concat layer of 13th normal cell as the feature for content reconstruction and the concat layer in first reduction cell as the feature for style reconstruction, the types of layers in each cell are shown in Fig.\ref{best-arch} and Fig.\ref{ap:3382}.

\subsection{Image captioning}
The training dataset of image captioning is MSCOCO\cite{LinMBHPRDZ14}, a large-scale dataset for the object detection, segmentation, and captioning. Each image is scaled to $224\times224$ in RGB channels and subtract the channel means as the input to a AlphaX-1 model. For training AlphaX-1 model, We use the SGD optimizer with the 16 batch size and the initial learning rate is 2.0. We applied the exponential learning rate decay schedule with the 0.5 decay factor in every 8 epochs.

\end{document}